%% file: arxiv.tex
\documentclass[10pt,twocolumn,letterpaper]{article}

\usepackage[pagenumbers]{cvpr} 

\usepackage{graphicx}
\usepackage{amsmath}
\usepackage{balance}
\usepackage{float}
\usepackage{booktabs}
\usepackage{booktabs}       
\usepackage{amsfonts}       
\usepackage{nicefrac}       
\usepackage{microtype}      
\usepackage[dvipsnames]{xcolor}
\usepackage{graphicx}
\usepackage{float}
\usepackage{makecell}
\usepackage{pifont}
\usepackage{tabu}
\usepackage{rotating}
\usepackage{enumitem}
\usepackage{natbib}
\usepackage{amssymb}
\usepackage{bm}
\usepackage{graphics}
\usepackage{multirow}
\usepackage{arydshln}
\usepackage{fixmath}
\usepackage{algorithm}
\usepackage{algpseudocode}
\usepackage{pifont}
\usepackage{mathtools}
\usepackage{xfrac}
\usepackage{colortbl,xcolor}
\usepackage{arydshln}
\usepackage{multirow}
\usepackage{xstring}
\usepackage{xcolor}
\input{preamble}

\definecolor{cvprblue}{rgb}{0.21,0.49,0.74}
\usepackage[pagebackref,breaklinks,colorlinks,allcolors=cvprblue]{hyperref}

\def\paperID{12190} 
\def\confName{CVPR}
\def\confYear{2026}

\title{BluRef: Unsupervised Image Deblurring with Dense-Matching References}

\newcommand{\psnr}[1]{{\textcolor{RedViolet}{#1}}}
\newcommand{\ssim}[1]{{\textcolor{Blue}{#1}}}

 \newcommand{\metric}[1]{%
     \StrBefore{#1}{/}[\numerator]%
     \StrBehind{#1}{/}[\denominator]%
     \textcolor{RedViolet}{\numerator }%
     \textcolor{black}{$\,$/$\,$}%
     \textcolor{Blue}{\denominator}%
 }

\newcommand{\metricu}[1]{%
    \StrBefore{#1}{/}[\numerator]%
    \StrBehind{#1}{/}[\denominator]%
    \textcolor{RedViolet}{\underline{\numerator}}%
    \textcolor{black}{$\,$/$\,$}%
    \textcolor{Blue}{\underline{\denominator}}%
}

\vspace{-15mm}
\author{
Bang-Dang Pham$^{1\dagger}$ \quad Anh Tran$^{2}$ \quad Cuong Pham$^{2,3}$ \quad Minh Hoai$^{2,4}$\\ 
\small{\textsuperscript{1}University of Wisconsin-Madison  \quad \textsuperscript{2}Qualcomm AI Research$^{\ddagger}$ \quad \textsuperscript{3}Posts \& Telecommunications Inst. of Tech. \quad \textsuperscript{4}Adelaide University}\\
\small{\url{https://qualcomm-ai-research.github.io/BluRef/}}
}
\setlist{nosep}
\makeatletter

\begin{document}
\input{definitions}

\twocolumn[{%
\renewcommand\twocolumn[1][]{#1}%
\maketitle
\begin{center}
    \vspace{-3mm}
    \includegraphics[width=0.98\linewidth]{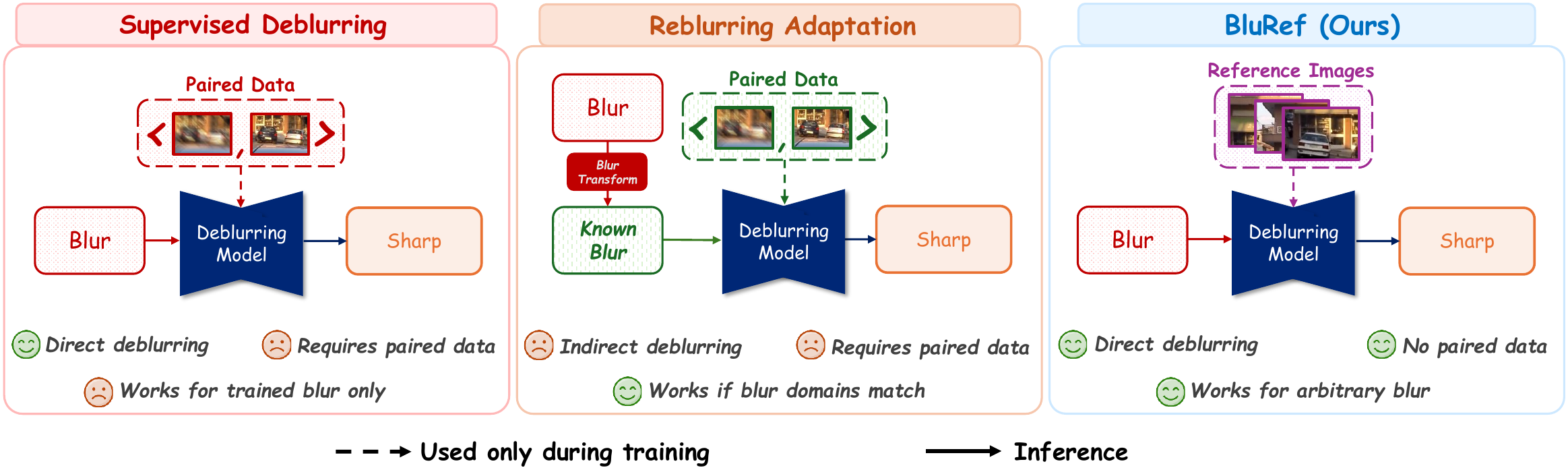}
    
    {\captionof{figure}{
Comparison of three approaches to image deblurring. Supervised deblurring requires costly paired data, limiting scalability. Reblurring-based methods rely on indirect mappings and suitable intermediate domains, which can be hard to identify. In contrast, the proposed BluRef directly learns from unpaired blurry and reference images within the target domain, offering an efficient and scalable solution without requiring paired supervision or complex multi-stage pipelines.}   
    \label{fig:teaser}}
    \vspace{-1mm}
\end{center}%
}]
\maketitle
\begingroup
\renewcommand\thefootnote{}
\footnotetext{\hspace{-1em}$^{\dagger}$Work done while at Qualcomm AI Research.}
\footnotetext{\hspace{-1em}$^{\ddagger}$Qualcomm AI Research is an initiative of Qualcomm Technologies, Inc.}
\endgroup

\vspace{-1mm}
\input{sec/0_abstract}    
\vspace{-6mm}
\input{sec/1_intro}
\input{sec/2_related}

\input{sec/3_method}

\input{sec/4_exp}

\input{sec/5_conclusion}
{
    \small
    \bibliographystyle{ieeenat_fullname}
    \bibliography{main}
}


\end{document}


\maketitle
\input{definitions}

\begin{abstract}
 In this supplementary material, we provide extended technical details and additional results supporting the BluRef framework. We first present implementation information and training-time analysis, clarifying how the full pipeline is executed in practice. We then describe the training procedure of the Dense Matching ($\mathcal{DM}$) module in greater depth, including the construction of synthetic training pairs, the degradation strategies used, and the role of augmentation in enabling robust correspondence learning. Additional ablation studies are included to further validate the design choices of BluRef, particularly regarding $\mathcal{DM}$ training configurations, reference aggregation strategies, and alternative deblurring backbones. We also report quantitative results for the Restormer backbone and provide extensive qualitative visualizations on GoPro, RB2V, and PhoneCraft, illustrating the improvements achieved by BluRef under diverse real-world blur conditions. Finally, we visualize the evolution of pseudo-sharp images and confidence masks across training iterations to highlight how the model progressively refines its supervision signals. All code and resources used in our experiments are included for reference.
\end{abstract}

\section{Extra System Details}
\subsection{Extra Implementation Detail}
During BluRef's pipeline training, each iteration processes an input image alongside a list of reference images, embedded into the deblurring model's backbone as data preprocessing. We utilize a batch size of 8, with code details provided in the supplementary materials.

\subsection{Training time} Embedding an iterative refinement technique by Truong \etal \cite{truong2023pdc}, our training period lasts 4 days on 1 NVIDIA A100 GPU. This duration is approximately 1.2 times longer than that of supervised models, a difference we consider negligible given the ability to train on an unpaired dataset and subsequently achieve pseudo-sharp images.

\subsection{Augmentation in DenseMatching Model}
\begin{figure}[t]
    \centering
        
    \includegraphics[width=\linewidth]{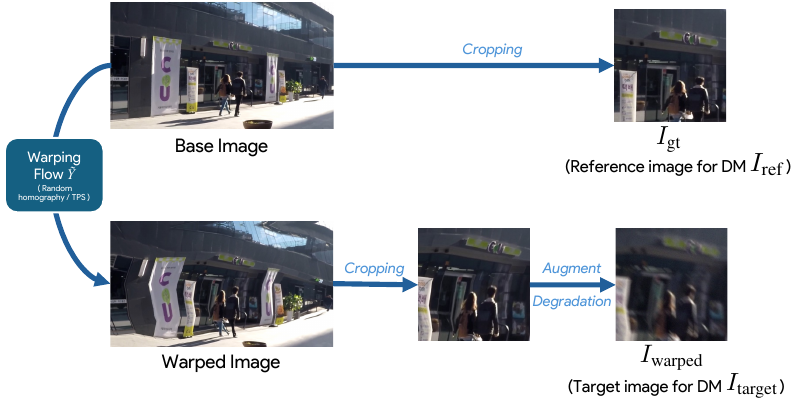}
    \caption{\textbf{Training pipeline of the dense matching module $\mathcal{DM}$}. A sharp image is resized and randomly warped to create a deformed view. Center crops from the original and deformed images form a synthetic pair $(I_{\text{warped}}, I_{\text{gt}})$, where $I_{\text{warped}}$ is further degraded using BSRGAN Augmentation which is fixed to add blur augmentation. The $\mathcal{DM}$ learns to map $(I_{\text{warped}}, I_{\text{gt}})$ to a transformed image $I_{\text{trans}}$ and a confidence mask $M_{\text{conf}}$.}
    \label{fig:dm_train_pipeline}
\end{figure}

As described in Sec.~\textcolor{red}{3.2} of the main paper and illustrated in our synthetic-pair generation pipeline (\cref{fig:dm_train_pipeline}), we adopt the BSRGAN degradation pipeline~\cite{zhang2021designing} to augment the synthetic training pairs $(I_{\text{warped}}, I_{\text{gt}})$ for the Dense Matching ($\mathcal{DM}$) model. This augmentation is essential for enabling the DM module to learn reliable correspondences across blurred and sharp domains. 

While BSRGAN was originally developed for blind image super-resolution, its diverse degradation space makes it well suited for our setting. We adapt the pipeline for blind image deblurring by expanding it with additional blur-oriented degradations. Concretely, our augmentation includes:
\begin{itemize}
    \item \textbf{Gaussian blur:} kernel sizes sampled from $\{7, 9, 11\}$ with random variance
    \item \textbf{Anisotropic and isotropic blur} following the standard BSRGAN blocks
    \item \textbf{Synthetic motion blur} implemented using directional/diagonal averaging kernels and a simple frame-averaging strategy to mimic real motion smear
    \item \textbf{Gaussian noise} with random variance
    \item \textbf{JPEG compression noise} with quality factor $q \sim U(30, 60)$
    \item \textbf{Random downsampling/upsampling} (bicubic or bilinear)
    \item \textbf{Degradation shuffle}, where all degradation operations are applied in a randomized order.
\end{itemize}

These augmentations generate a wide spectrum of blur and noise patterns, including realistic motion smear produced by averaging, enabling the $\mathcal{DM}$ model to learn dense correspondences under heterogeneous and challenging blur conditions. Importantly, keeping the degradation shuffle ensures the model is trained on a richly varied distribution of warped images, improving its robustness when matching real-world blurry inputs.

\section{Restormer's Quantitative Result}
As mentioned in the main paper, we provide the quantitative results of Restormer using the Weighted Average (Avg.) and Sequential Accumulation (Seq.) strategies. Additional details are available in \Tref{tab:quan1}.

\section{Additional Ablation Studies}
In the main paper, we conducted extensive ablation experiments to assess the effects of the number of reference frames, the pseudo-ground truth generation module, varying $\Delta$
values, different deblurring backbones, and diverse refinement strategies. This supplementary material presents additional ablation studies detailed below.

\subsection{BluRef with Different DenseMatching Models}
In the main paper, we employ PDC-Net+ as the DM to train BluRef, leveraging its state-of-the-art performance over other baselines. To assess the impact of different DM models, we substitute PDC-Net+ with a pre-trained GLU-Net \cite{truong2020glu}, trained on the same paired $(I_{\text{warped}}, I_{\text{gt}})$ set as PDC-Net+, and follow the identical BluRef training pipeline. Using NAFNet on RB2V ($\Delta = 10$) with the Progressive (Prog.) strategy. \cref{tab:dm} shows that PSNR$\uparrow$/SSIM$\uparrow$ reaches 26.43/0.792, which is comparable to PDC-Net+ and approaches the upperbound model. This result highlights the robustness of our BluRef pipeline.
\begin{table}[t]
    \centering
    \aboverulesep=0ex 
    \belowrulesep=0ex 
    \begin{tabular}{l|cc}
        \toprule
         Models            & PSNR$\uparrow$ & SSIM$\uparrow$\\
        \midrule
        BluRef + PDC-Net+  & 27.72 & 0.820\\
        BluRef + GLU-Net  & 26.63 & 0.798  \\
        Upperbound (NAFNet) & 28.54 & 0.824 \\
        \bottomrule
    \end{tabular}
    \caption{Comparison of BluRef training between with PDC-Net+ and GLU-Net. \label{tab:dm}}
\end{table}

\subsection{BluRef with GAN-Based Deblurring Backbones}
 To validate BluRef’s effectiveness with other GAN-based deblurring backbones, we integrate UID-GAN—an unsupervised deblurring backbone benchmarked in the main paper—into our BluRef framework, employing the Progressive (Prog.) strategy with the same DM model (PDC-Net+) and settings on RB2V ($\Delta = 10$). To adapt its unsupervised training to BluRef, we apply the same reconstruction loss as used with NAFNet. As can be seen in \cref{tab:gan}, this boosted PSNR$\uparrow$/SSIM$\uparrow$ to 24.56/0.698, compared to 22.29/0.581 for the original UID-GAN, confirming BluRef’s robust performance across diverse deblurring backbones. 
\begin{table}[t]
    \centering
    \aboverulesep=0ex 
    \belowrulesep=0ex 
    \begin{tabular}{l|cc}
        \toprule
         Models            & PSNR$\uparrow$ & SSIM$\uparrow$\\
        \midrule
        UID-GAN  & 22.01 & 0.551 \\
        UID-GAN + BluRef & 24.56 & 0.698 \\
        \bottomrule
    \end{tabular}
    \caption{Ablation studies with UID-GAN deblurring backbone. \label{tab:gan}}
\end{table}

\subsection{Effect of DenseMatching training setting on BluRef Performance}
\begin{table}[t]
\centering
\aboverulesep=0pt
\belowrulesep=0pt
\begin{tabular}{l|c|c}
\toprule
$\mathcal{DM}$ Training Setting & PSNR$\uparrow$ & SSIM$\uparrow$ \\
\midrule
No augmentation & 27.52 & 0.802 \\
Gaussian blur only & 29.05 & 0.861 \\
Gaussian + Noise & 29.21 & 0.868 \\
Motion blur only & 29.44 & 0.874 \\
Down/Up sampling only & 28.93 & 0.855 \\
\midrule
BSRGAN (original) & 30.58 & 0.892 \\
\textbf{BSRGAN + Motion Blur (Ours)} &  \textbf{31.87} & \textbf{0.955 } \\
\bottomrule
\end{tabular}
\caption{\textbf{Ablation study of the Dense Matching (DM) training pipeline.} 
We compare different degradation configurations used to synthesize $(I_{\text{warped}}, I_{\text{gt}})$ training pairs. 
Results are reported on the GoPro validation set.}
\label{tab:dm_ablation}
\end{table}

\begin{figure*}[!htbp]
    \centering
    \includegraphics[width=\textwidth]{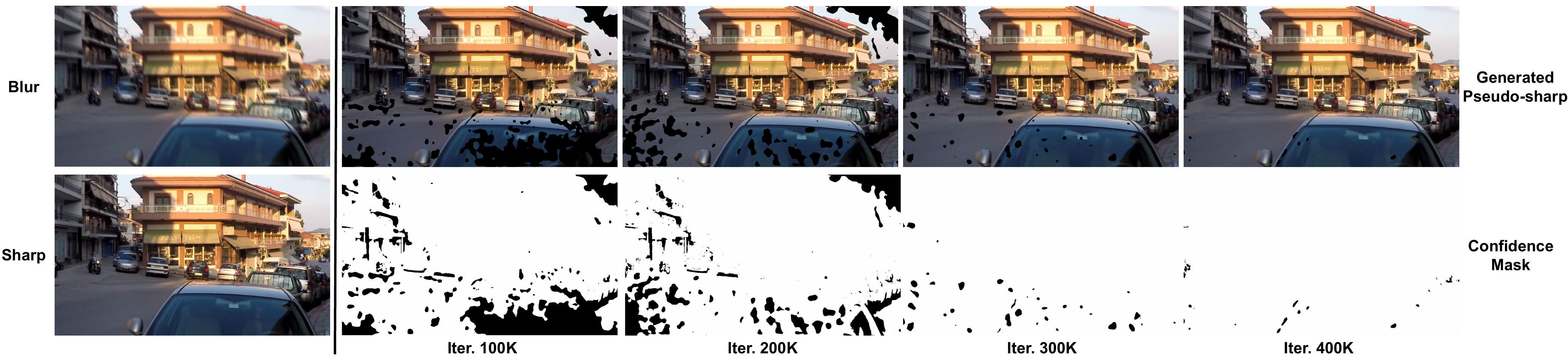}
    \caption{Our generated pseudo-sharp images and their confidence masks follow by iterations.}
    \label{fig:pseudo}
\end{figure*}

\begin{figure*}[!htb]
    \centering
    \includegraphics[width=\textwidth]{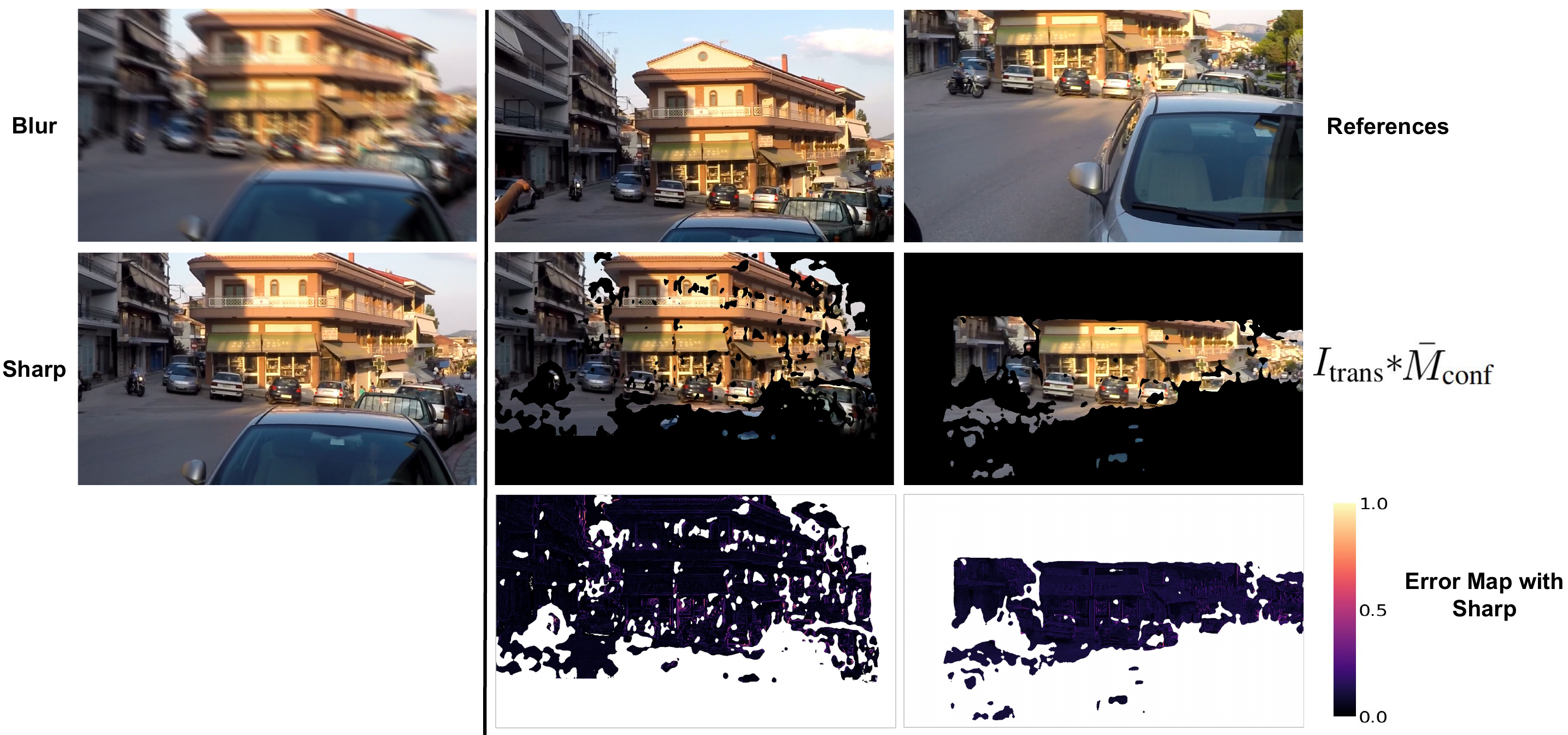}
     \caption{\textbf{Visualization of the Dense Matching output $I_{\text{trans}}$ at the first iteration.}
    The top row shows the blurry input and three sharp reference frames.
    The middle row displays the ground-truth sharp image and the DM output
    $I_{\text{trans}} * \bar{M}_{\text{conf}}$, where $\bar{M}_{\text{conf}}$ is the binary confidence mask.
    The bottom row shows the pixel-wise error map between $I_{\text{trans}}$ and the sharp image.
    Despite being generated at the first iteration and without any refinement, $I_{\text{trans}}$ already
    captures high-frequency scene structures from the references, demonstrating that the Dense Matching
    module successfully transfers sharp details before the pseudo-sharp aggregation stage (Zoom in for best view).}
    \label{fig:dm_error}
\end{figure*}

This ablation examines how different degradation settings used to train the 
Dense Matching (DM) model influence the final BluRef performance. 
For each degradation configuration, we train a separate DM model on synthetic
$(I_{\text{warped}}, I_{\text{gt}})$ pairs generated by geometric warping 
, followed by one of the following degradation pipelines:
(i)~No blur or noise (identity), 
(ii)~Gaussian blur only, 
(iii)~Gaussian blur + noise, 
(iv)~Motion blur only, 
(v)~Downsampling/upsampling, 
(vi)~BSRGAN degradations, and 
(vii)~BSRGAN degradations with additional motion blur (Ours).  
We then keep the entire BluRef framework fixed (NAFNet backbone, 
$\Delta{=}10$, $N{=}6$ , Progressive Reference Strategy) and plug in each DM model to generate pseudo-sharp
targets and train the deblurring network.  
Thus, the PSNR/SSIM values in \cref{tab:dm_ablation} directly reflect 
how each DM training variant affects the final BluRef performance on the GoPro dataset.

The results indicate that training DM without blur augmentation leads to a significant drop in BluRef performance, showing that the matching module fails to generalize to real blurry inputs. Simple degradations (Gaussian, noise, motion blur, or downsampling) offer only  moderate improvements and remain insufficient to bridge the synthetic--real gap.   The original BSRGAN pipeline already yields a strong gain, thanks to its randomized degradation shuffle that exposes the DM model to diverse distortions. Finally, BSRGAN + Motion Blur (Ours) achieves the best PSNR/SSIM, as the additional motion-averaging blur improves robustness to camera shake and produces more stable correspondences. This confirms that a rich combination of geometric warps and different degradations is essential for training a DM model that can generalize to the diverse, unconstrained distortions present in real-world blurry images.

\subsection{Impact of Pseduo ground truth generation} 
To study the necessity of pseudo-ground truth aggregation, we compare the performance of BluRef with and without aggregation strategies for generating pseudo-ground truth $I_{\text{pseudo}}$. Specifically, we train BluRef on RB2V and GoPro ($\Delta = 10$) using NAFNet as the backbone, but without aggregation strategies, forcing the model to learn from individual supervision signals $I_{\text{pseudo}}^n$ instead of the aggregated supervision signal $I_{\text{pseudo}}$ as proposed. As shown in \cref{tab:strategy}, this leads to a significant performance drop due to inconsistent supervision signal during deblurring model training, underscoring the crucial role of our iterative refinement with reference images.

\begin{table}[t]
    \centering
    \aboverulesep=0ex 
    \belowrulesep=0ex 
    \resizebox{0.98\columnwidth}{!}{
    \setlength{\tabcolsep}{5pt}
    \begin{tabular}{l|cc}
        \toprule
         Dataset            & GoPro & RB2V \\
        \midrule
        BluRef (Prog.) & \textbf{\metric{31.87/0.955}} & \textbf{\metric{27.72/0.820 }}\\
        BluRef w/o aggregation strategy & \metric{18.89/0.323} & \metric{18.73/0.325}  \\
        \bottomrule
    \end{tabular}
    }
    \vskip -0.05in
    \caption{Comparison of BluRef training with and without the aggregation strategy. For each test, we report \psnr{PSNR$\uparrow$}/\ssim{SSIM$\uparrow$} scores as evaluation metrics.   \label{tab:strategy}}
    \vspace{-4mm}
\end{table}

\section{Additional Visualization}
\subsection{Qualitative Results in GoPro and RB2V dataset}
\begin{figure*}[!htb]
    \centering
    \includegraphics[width=0.8\textwidth]{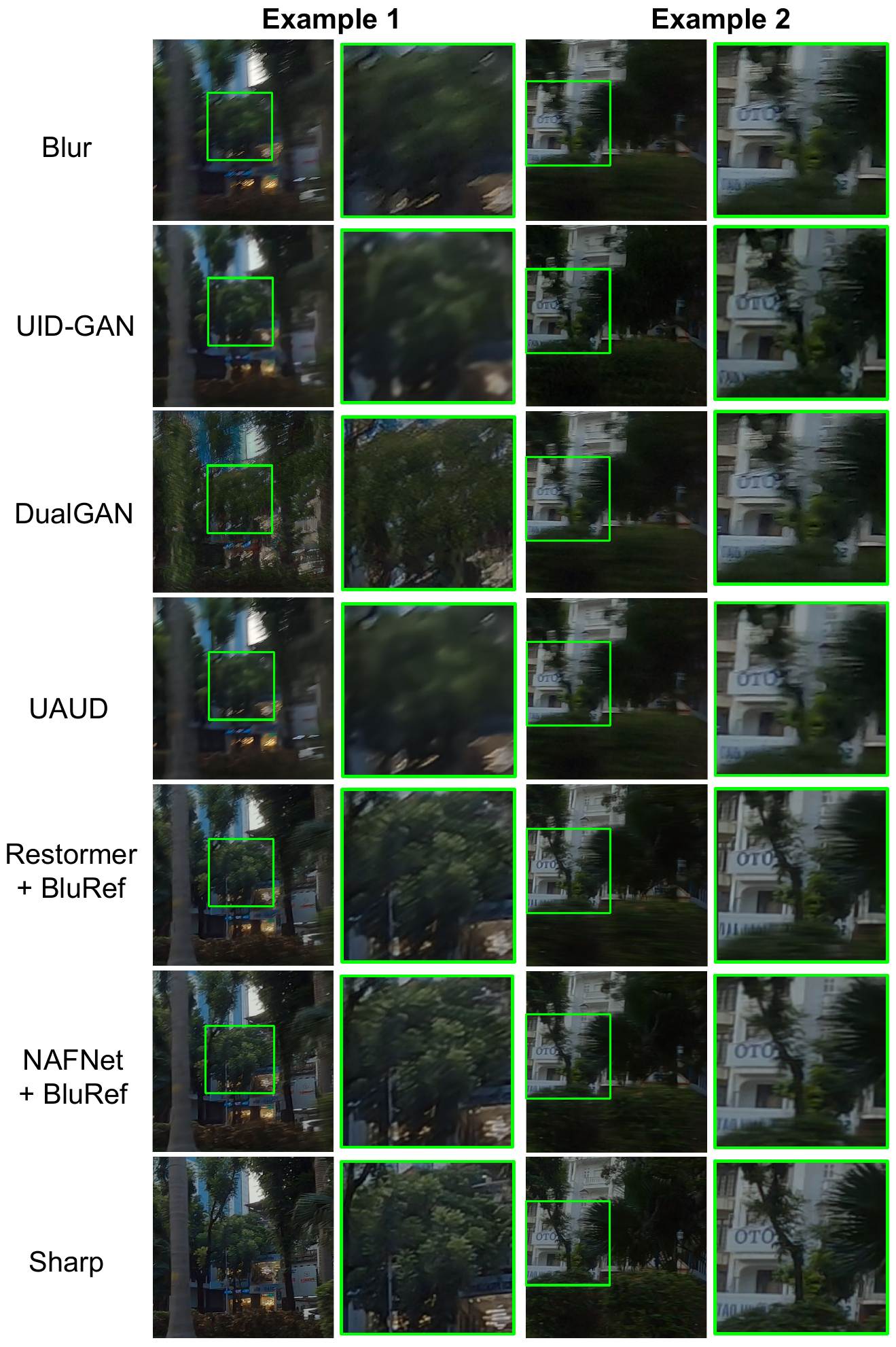}
    \caption{Addtional Qualitative Results in RB2V Dataset.}
    \label{fig:rb2v}
\end{figure*}

\begin{figure*}[!htb]
    \centering
    \includegraphics[width=\textwidth]{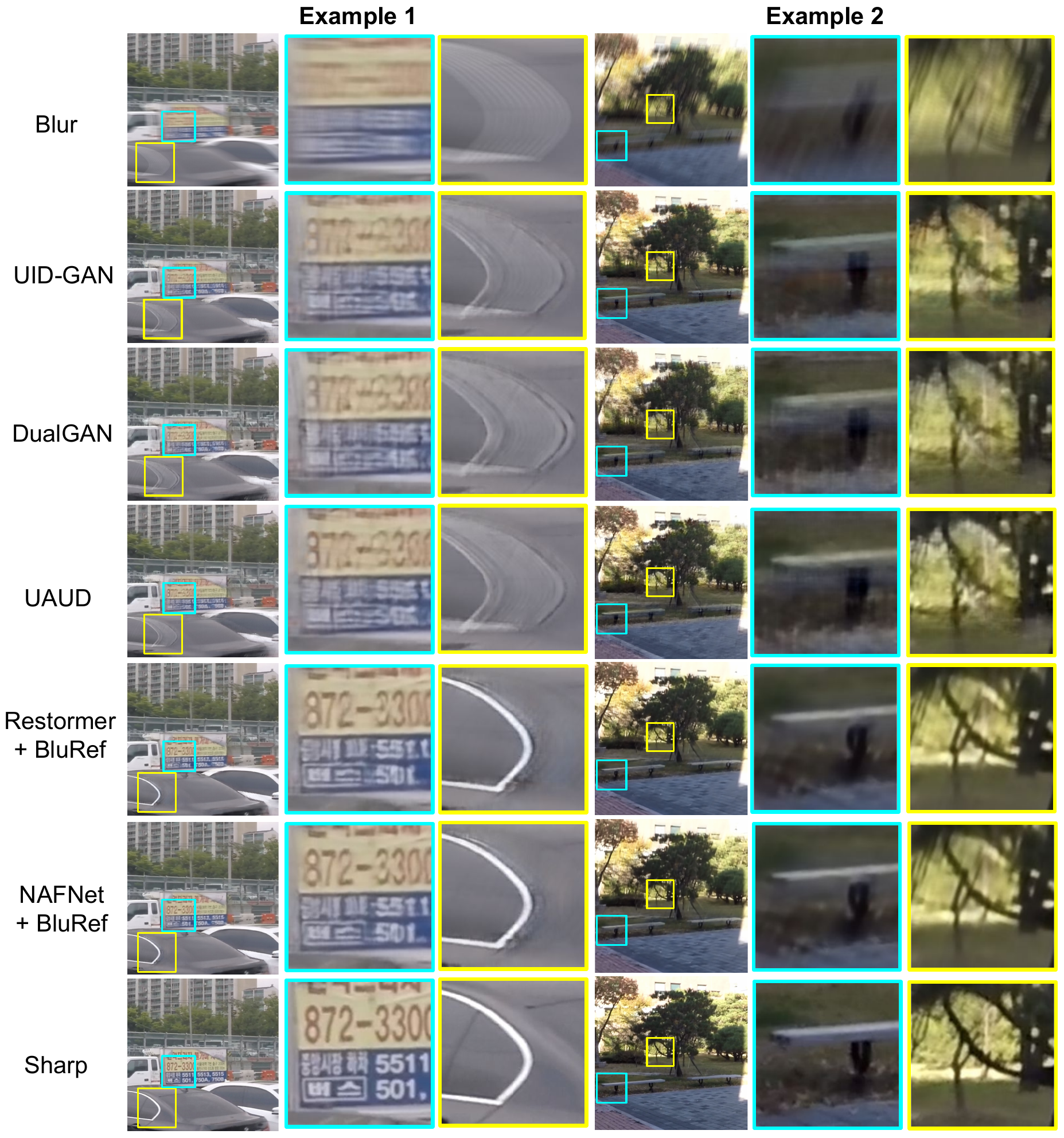}
    \caption{Addtional Qualitative Results in GoPro Dataset.}
    \label{fig:gopro}
\end{figure*}

In this section, we provide additional qualitative figures comparing the image deblurring results of our BluRef and other baselines. Figures \ref{fig:rb2v} and  \ref{fig:gopro} provide samples on GoPro\cite{gopro} and RB2V\cite{pham2023hypercut}, respectively.

\subsection{Qualitative Results in PhoneCraft dataset} 
\begin{figure*}[!htb]
    \centering
    \includegraphics[width=\textwidth]{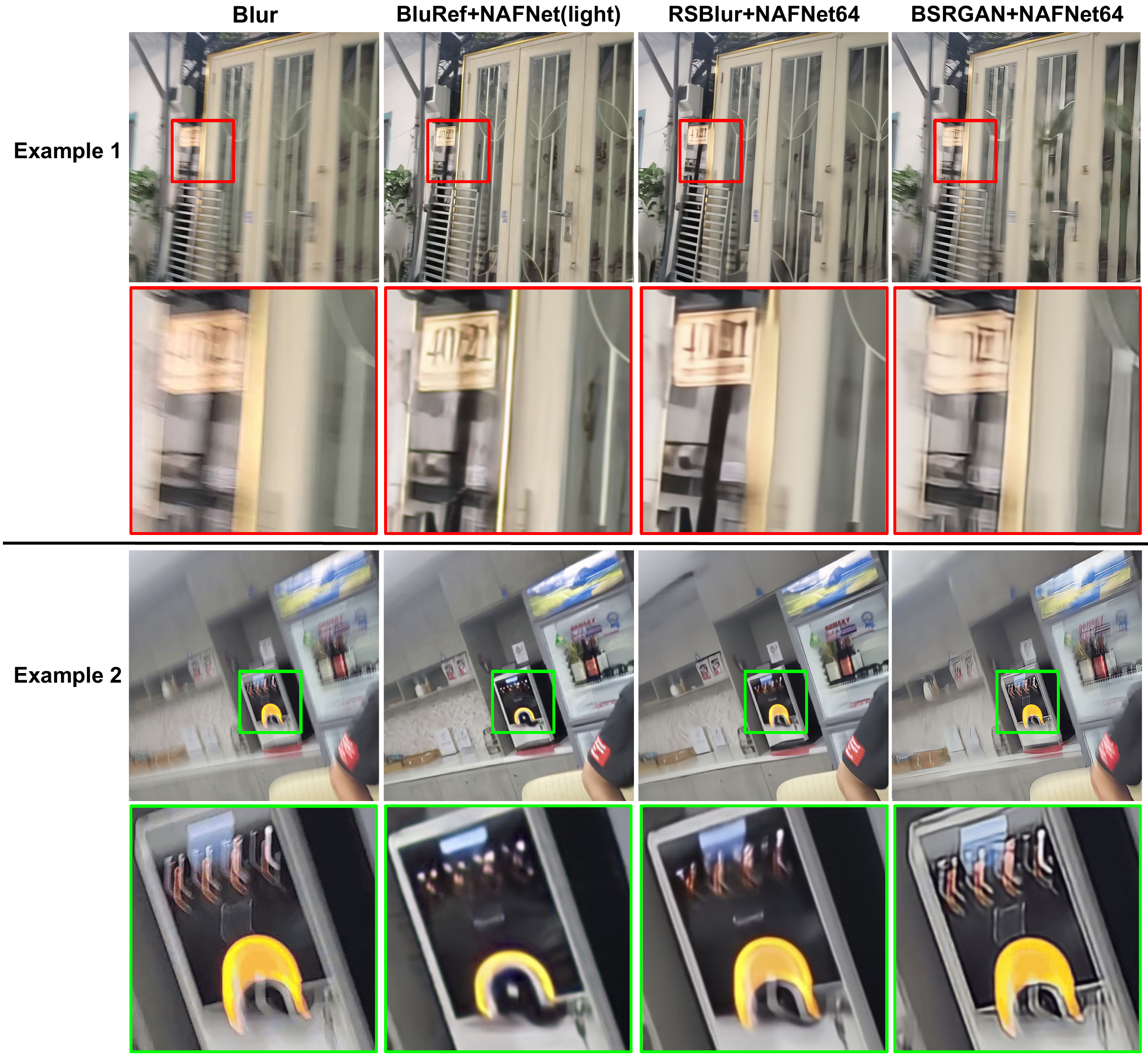}
    \caption{The Qualitative Results in PhoneCraft dataset.}
    \label{fig:phonecraft}
\end{figure*}

As discussed in Sec. \textcolor{red}{4.4} - Evaluating on ‘BluRef dataset’ of the main manuscript, we validate our generated pseudo-sharp images by pairing them with corresponding blurry images to form the paired version of PhoneCraft\cite{pham2024blur2blur} dataset. We train a compact NAFNet on this dataset for performance benchmarking against generalized deblurring models BSRGAN and RSBlur, both utilizing NAFNet64. We report the qualitative results of these comparisons in \cref{fig:phonecraft}. As shown in \cref{fig:phonecraft}, training on our pseudo-sharp images facilitates the lightweight version of NAFNet to achieve sufficient and superior performance compared to BSRGAN and RSBlur, which use the default version of NAFNet. These examples demonstrate that our `BluRef dataset' effectively enables the supervised network capture the blur kernel of real-world blur, reconstruct sharp details accurately, and avoid artifacts or excessive smoothing.

\begin{table*}[!htbp]

\centering
\aboverulesep=0ex 
\belowrulesep=0ex 

\resizebox{\linewidth}{!}{%
\setlength{\tabcolsep}{5pt}
\begin{tabular}{l|ccc|ccc}
\toprule\rule{0pt}{1.1EM}
             & \multicolumn{3}{c|}{\textbf{GoPro}}    & \multicolumn{3}{c}{\textbf{RB2V}} \\
\midrule\rule{0pt}{1.1EM}
\textbf{Delta ($\Delta$)}         & \textit{1 frame}       & \textit{10 frames}      & \textit{20 frames}       & \textit{1 frame}       & \textit{10 frames}       & \textit{20 frames}     \\ 
\midrule
\rule{0pt}{1.1EM}
\quad \textbf{BluRef (Ours)} & & & & & & \\

\quad \quad Restormer - BluRef (Avg.)  & \metric{27.12/0.905} & \metric{27.04/0.895} & \metric{26.98/0.893} & \metric{25.41/0.816} & \metric{25.38/0.812} & \metric{24.78/0.801}  \\
\quad \quad Restormer - BluRef (Seq.) & \metric{28.46/0.923} & \metric{28.37/0.920} & \metric{28.31/0.912} & \metric{25.22/0.810} & \metric{25.20/0.811} & \metric{24.73/0.792}  \\
\midrule
\textit{\textbf{Supervised - Upperbound}} & & & & & & \\
\quad NAFNet     & \multicolumn{3}{c|}{\metric{33.32/0.962}} & \multicolumn{3}{c}{\metric{28.54/0.824}}  \\
\quad Restormer  & \multicolumn{3}{c|}{\metric{32.92/0.961}} & \multicolumn{3}{c}{\metric{27.43/0.849}}  \\
\bottomrule
\end{tabular}
}
\caption{Comparison of our proposed BluRef (Restormer backbone) on GoPro and RB2V datasets. For each test, we report \psnr{PSNR$\uparrow$}/\ssim{SSIM$\uparrow$} scores as evaluation metrics.} \label{tab:quan1}
\end{table*}


\subsection{Visualization of Pseudo-sharp images}
In this section, we present the evolution of pseudo-sharp images generated by our model through training iterations with a given blurry input. This experiment was conducted using the GoPro dataset, with $\Delta = 10$ and 6 reference frames utilized for training. As illustrated in \cref{fig:pseudo}, the quality of the pseudo-sharp images is enhanced progressively with increased iterations. Notably, at the final iteration of 400K, the pseudo-sharp image exhibits a high degree of similarity to the actual ground truth image, confirming the effectiveness of our model in synthesizing sharp images from blurred inputs.

Despite the high fidelity of the generated images, there is a residual impact from the confidence masks applied during the image restoration process. As can be seen in the \cref{fig:pseudo}, the black regions in confidence mask identify regions with lower reliability, have a limited influence on the learning of the blur kernel space. However, this influence is marginal and does not substantially degrade the deblurring performance, as evidenced by the minimal deviation from the ground truth in the final iteration. This underscores the robustness of our approach in handling artifacts by efficient iterative refinement process.

\myheading{Visualization of $\mathcal{DM}$ results before aggregation.}To analyze the behavior of the $\mathcal{DM}$ module, we visualize its transformed output
$I_{\text{trans}}$ at the very first iteration of BluRef training.
As shown in \cref{fig:dm_error}, the $\mathcal{DM}$ output, when masked by its confidence map, already reconstructs meaningful edges, building contours, and scene geometry that closely resemble the ground-truth sharp image.
While the reconstruction is still imperfect and contains
notable mismatches, the corresponding error map shows that a substantial portion of the scene already
exhibits relatively low error with respect to the sharp image.

This indicates that the DM module is able to transfer a meaningful amount of sharp structural information
from the reference frames even before any refinement or aggregation takes place. However, the incomplete
and spatially inconsistent regions also highlight the necessity of our iterative pseudo–ground-truth
aggregation strategy, which progressively stabilizes and improves the supervision signals used to train the deblurring network (see the gradual improvements across iterations in \cref{fig:pseudo}).


    
    

{
    \small
    \bibliographystyle{ieeenat_fullname}
    \bibliography{main}
}


%% file: preamble.tex
\usepackage{xstring}
\usepackage{lineno}
\usepackage{capt-of}



%% file: definitions.tex
\def\mA{\mathcal{A}}
\def\mB{\mathcal{B}}
\def\mC{\mathcal{C}}
\def\mD{\mathcal{D}}
\def\mE{\mathcal{E}}
\def\mF{\mathcal{F}}
\def\mG{\mathcal{G}}
\def\mH{\mathcal{H}}
\def\mI{\mathcal{I}}
\def\mJ{\mathcal{J}}
\def\mK{\mathcal{K}}
\def\mL{\mathcal{L}}
\def\mM{\mathcal{M}}
\def\mN{\mathcal{N}}
\def\mO{\mathcal{O}}
\def\mP{\mathcal{P}}
\def\mQ{\mathcal{Q}}
\def\mR{\mathcal{R}}
\def\mS{\mathcal{S}}
\def\mT{\mathcal{T}}
\def\mU{\mathcal{U}}
\def\mV{\mathcal{V}}
\def\mW{\mathcal{W}}
\def\mX{\mathcal{X}}
\def\mY{\mathcal{Y}}
\def\mZ{\mathcal{Z}} 

\def\bbN{\mathbb{N}} 
\def\bbR{\mathbb{R}} 
\def\bbP{\mathbb{P}} 
\def\bbQ{\mathbb{Q}} 
\def\bbE{\mathbb{E}}

\def\1n{\mathbf{1}_n}
\def\0{\mathbf{0}}
\def\1{\mathbf{1}}

\def\A{{\bf A}}
\def\B{{\bf B}}
\def\C{{\bf C}}
\def\D{{\bf D}}
\def\E{{\bf E}}
\def\F{{\bf F}}
\def\G{{\bf G}}
\def\H{{\bf H}}
\def\I{{\bf I}}
\def\J{{\bf J}}
\def\K{{\bf K}}
\def\L{{\bf L}}
\def\M{{\bf M}}
\def\N{{\bf N}}
\def\O{{\bf O}}
\def\P{{\bf P}}
\def\Q{{\bf Q}}
\def\R{{\bf R}}
\def\S{{\bf S}}
\def\T{{\bf T}}
\def\U{{\bf U}}
\def\V{{\bf V}}
\def\W{{\bf W}}
\def\X{{\bf X}}
\def\Y{{\bf Y}}
\def\Z{{\bf Z}}

\def\a{{\bf a}}
\def\b{{\bf b}}
\def\c{{\bf c}}
\def\d{{\bf d}}
\def\e{{\bf e}}
\def\f{{\bf f}}
\def\g{{\bf g}}
\def\h{{\bf h}}
\def\i{{\bf i}}
\def\j{{\bf j}}
\def\k{{\bf k}}
\def\l{{\bf l}}
\def\m{{\bf m}}
\def\n{{\bf n}}
\def\o{{\bf o}}
\def\p{{\bf p}}
\def\q{{\bf q}}
\def\r{{\bf r}}
\def\s{{\bf s}}
\def\t{{\bf t}}
\def\u{{\bf u}}
\def\v{{\bf v}}
\def\w{{\bf w}}
\def\x{{\bf x}}
\def\y{{\bf y}}
\def\z{{\bf z}}

\def\balpha{\mbox{\boldmath{$\alpha$}}}
\def\bbeta{\mbox{\boldmath{$\beta$}}}
\def\bdelta{\mbox{\boldmath{$\delta$}}}
\def\bgamma{\mbox{\boldmath{$\gamma$}}}
\def\blambda{\mbox{\boldmath{$\lambda$}}}
\def\bsigma{\mbox{\boldmath{$\sigma$}}}
\def\btheta{\mbox{\boldmath{$\theta$}}}
\def\bomega{\mbox{\boldmath{$\omega$}}}
\def\bxi{\mbox{\boldmath{$\xi$}}}
\def\bnu{\mbox{\boldmath{$\nu$}}}                                  
\def\bphi{\mbox{\boldmath{$\phi$}}}
\def\bmu{\mbox{\boldmath{$\mu$}}}

\def\bDelta{\mbox{\boldmath{$\Delta$}}}
\def\bOmega{\mbox{\boldmath{$\Omega$}}}
\def\bPhi{\mbox{\boldmath{$\Phi$}}}
\def\bLambda{\mbox{\boldmath{$\Lambda$}}}
\def\bSigma{\mbox{\boldmath{$\Sigma$}}}
\def\bGamma{\mbox{\boldmath{$\Gamma$}}}
                                  
\newcommand{\myprob}[1]{\mathop{\mathbb{P}}_{#1}}

\newcommand{\myexp}[1]{\mathop{\mathbb{E}}_{#1}}

\newcommand{\mydelta}[1]{1_{#1}}

\newcommand{\myminimum}[1]{\mathop{\textrm{minimum}}_{#1}}
\newcommand{\mymaximum}[1]{\mathop{\textrm{maximum}}_{#1}}    
\newcommand{\mymin}[1]{\mathop{\textrm{minimize}}_{#1}}
\newcommand{\mymax}[1]{\mathop{\textrm{maximize}}_{#1}}
\newcommand{\mymins}[1]{\mathop{\textrm{min.}}_{#1}}
\newcommand{\mymaxs}[1]{\mathop{\textrm{max.}}_{#1}}  
\newcommand{\myargmin}[1]{\mathop{\textrm{argmin}}_{#1}} 
\newcommand{\myargmax}[1]{\mathop{\textrm{argmax}}_{#1}} 
\newcommand{\myst}{\textrm{s.t. }}

\newcommand{\denselist}{\itemsep -1pt}
\newcommand{\sparselist}{\itemsep 1pt}

\definecolor{pink}{rgb}{0.9,0.5,0.5}
\definecolor{purple}{rgb}{0.5, 0.4, 0.8}   
\definecolor{gray}{rgb}{0.3, 0.3, 0.3}
\definecolor{mygreen}{rgb}{0.2, 0.6, 0.2}

\newcommand{\cyan}[1]{\textcolor{cyan}{#1}}
\newcommand{\blue}[1]{\textcolor{blue}{#1}}
\newcommand{\magenta}[1]{\textcolor{magenta}{#1}}
\newcommand{\pink}[1]{\textcolor{pink}{#1}}
\newcommand{\green}[1]{\textcolor{green}{#1}} 
\newcommand{\gray}[1]{\textcolor{gray}{#1}}    
\newcommand{\mygreen}[1]{\textcolor{mygreen}{#1}}    
\newcommand{\purple}[1]{\textcolor{purple}{#1}}       

\definecolor{greena}{rgb}{0.4, 0.5, 0.1}
\newcommand{\greena}[1]{\textcolor{greena}{#1}}

\definecolor{bluea}{rgb}{0, 0.4, 0.6}
\newcommand{\bluea}[1]{\textcolor{bluea}{#1}}
\definecolor{reda}{rgb}{0.6, 0.2, 0.1}
\newcommand{\reda}[1]{\textcolor{reda}{#1}}

\def\changemargin#1#2{\list{}{\rightmargin#2\leftmargin#1}\item[]}
\let\endchangemargin=\endlist
                                               
\newcommand{\cm}[1]{}

\newcommand{\mhoai}[1]{{\color{magenta}\textbf{[MH: #1]}}}

\newcommand{\mtodo}[1]{{\color{red}$\blacksquare$\textbf{[TODO: #1]}}}
\newcommand{\myheading}[1]{\vspace{0.5ex}\noindent \textbf{#1}}
\newcommand{\htimesw}[2]{\mbox{$#1$$\times$$#2$}}


%
%
%

\newcommand{\Sref}[1]{Sec.~\ref{#1}}
\newcommand{\Eref}[1]{Eq.~(\ref{#1})}
\newcommand{\Fref}[1]{Fig.~\ref{#1}}
\newcommand{\Tref}[1]{Table~\ref{#1}}

%% file: sec/0_abstract.tex
\begin{abstract}

This paper introduces a novel unsupervised approach for image deblurring that utilizes a simple process for training data collection, thereby enhancing the applicability and effectiveness of deblurring methods. Our technique does not require meticulously paired data of blurred and corresponding sharp images; instead, it uses unpaired blurred and sharp images of similar scenes to generate pseudo-ground truth data by leveraging a dense matching model to identify correspondences between a blurry image and reference sharp images. Thanks to the simplicity of the training data collection process, our approach does not rely on existing paired training data or pre-trained networks, making it more adaptable to various scenarios and suitable for networks of different sizes, including those designed for low-resource devices. We demonstrate that this novel approach achieves state-of-the-art performance, marking a significant advancement in the field of image deblurring.
\end{abstract}

%% file: sec/1_intro.tex
\section{Introduction}
\label{sec:intro}

Motion blur is a common issue in images and videos, reducing their visual quality and potentially impairing the performance of subsequent computer vision applications. Therefore, a reliable image deblurring technique is crucial in many scenarios. Currently, the leading approach is data-driven, involving the training of a deblurring function using paired data comprising blurry and corresponding sharp images. However, acquiring such paired training data requires a complex setup, including a beam splitter, cameras of the same make operating at different speeds, and systems for time synchronization, geometric alignment, and color calibration \cite{rsblur, levin2009understanding, liang2020raw}. Most cameras are not equipped to meet these stringent requirements, and for those that are, the setup is cumbersome and can be nearly impossible in specific contexts, such as for a moving dashcam or bodycam. 
As a result, if the images or target domain come from scenarios where collecting paired training data is impractical—which is often the case—supervised deblurring methods become unusable.


One approach that circumvents the need for paired training data in the target blur domain is the reblurring-based method \cite{pham2024blur2blur, wu2024id}. Instead of directly deblurring the image, it first maps the blurry input to an intermediate blur domain where a supervised deblurring model is already available, and then applies that model to recover the sharp image. This blur-to-blur-to-sharp strategy can be effective if the intermediate domain closely resembles the target blur domain, since a mapping between the two blur domains must also be learned. However, identifying such a suitable intermediate domain is often challenging, and any mismatch can negatively impact performance. Furthermore, the multi-stage pipeline introduces additional computational overhead, making it less efficient than direct deblurring approaches.

Both of the aforementioned approaches share a fundamental limitation: the domain gap. This gap arises either from a mismatch between the target blur domain and the domain used to train the deblurring model, or from the challenge of identifying an intermediate domain sufficiently similar for effective mapping. This raises a natural question: \textit{Is it possible to eliminate the domain gap altogether?}

The logical place to find data without a domain gap is the target domain itself—using images drawn directly from the same distribution as the blurry images we aim to deblur. Such data is often readily available in large quantities. The main obstacle, however, is that these images typically lack paired sharp counterparts. But do we really need paired data to train a deblurring model? Our answer is no. This insight forms the core motivation of our work: we demonstrate that it is possible to train high-performing deblurring models entirely within the target domain by leveraging unpaired reference images and generating pseudo-ground truth supervision - without ever requiring paired blur/sharp datasets.


Specifically, we propose BluRef, a novel unpaired reference-guided deblurring framework that operates entirely within the target domain. BluRef learns from unpaired blurry and sharp reference images that do not need to be geometrically aligned or captured simultaneously. It leverages a Dense Matching ($\mathcal{DM}$) model, pretrained on synthetically warped data, to establish correspondences between the blurred input and sharp references, generating high-quality pseudo-ground-truth images. These pseudo-sharp targets are iteratively refined and used to supervise the deblurring network, eliminating the need for paired datasets or pretrained deblurring models. This design simplifies data collection—requiring only unpaired video frames—and confines the use of synthetic data to the $\mathcal{DM}$ pretraining stage, ensuring strong performance on real-world blur.

A valuable byproduct of BluRef is the generation of pseudo-ground-truth pairs during its iterative refinement process. These pairs can be reused to train deblurring networks of varying capacities, including lightweight models suitable for mobile or resource-constrained devices. This extends the practical utility of BluRef beyond a single backbone, enabling flexible deployment across a range of platforms. We further analyze the pseudo-pair generation mechanism and its downstream benefits in \cref{sec:benefits}.


The name BluRef reflects our use of reference images, but our approach should not be confused with reference-based deblurring methods \cite{liu2023reference, zou2023reference, zhao2022d2hnet}, which remain within the supervised learning paradigm. These methods require paired blur-sharp data during training and incorporate sharp references to enhance texture preservation during both training and inference. For example, \citet{lai2022face} employ a dual-camera setup with spatially aligned reference frames for blurry face enhancement. While such methods demonstrate the value of reference images, they do not address the core challenge of learning from unpaired, unconstrained data—central to our unsupervised BluRef framework.

In short, this paper makes three key contributions. {\bf First,} we introduce BluRef, the first unpaired reference-guided framework for unsupervised image deblurring,  eliminating the need for ground-truth supervision or paired synthetic data through iterative dense matching with unpaired references. {\bf Second,} we show that BluRef’s pseudo–ground-truth images can be reused to train models of varying capacities, including lightweight networks suitable for mobile deployment. {\bf Third,} extensive experiments indicate that BluRef outperforms existing unsupervised methods and approaches the quality of supervised models on both synthetic and real-world datasets.

%% file: sec/2_related.tex
\vspace{-2mm}
\section{Related Work}
\vspace{-1mm}
\label{sec:related}



Related to our image deblurring method are studies on reference-based image restoration and dense matching, which will be reviewed in this section.

\myheading{Image deblurring}. The proposed method stands out from other unsupervised deblurring techniques \cite{lu2019uid, yi2017dualgan, zhang2023neural, chen2024unsupervised} given its pioneering approach of using reference images to generate pseudo-ground truths for directly training a deblur network. A notable related approach, \citet{pham2024blur2blur}, converts unknown blur types into a known domain, allowing pre-trained models to handle previously unsupported blur types. However, unlike their method, which depends on a pre-trained network trained with paired data, our approach operates without this requirement. This independence is essential, as it removes the need to adapt the input blurry image to align with the capabilities of a pre-trained network, a process that can be challenging when the target deblurring domain differs significantly from those with available paired data---especially given the rigorous requirements for collecting such data. 
Recent supervised methods, such as ID-BLAU~\cite{wu2024id} and SAM-Deblur~\cite{li2024sam}, achieve strong performance but remain constrained by the need for paired datasets. In contrast, our method attains competitive performance in a fully unsupervised setting while maintaining strong adaptability across diverse blur domains. Furthermore, it surpasses unsupervised techniques that depend on synthetic data augmentation \cite{zhang2021designing, rsblur}, whose artificial blur generation often fails to represent real-world degradation, leading to limited generalization.

\myheading{Reference-based image restoration.}
Reference-based image restoration, particularly in super-resolution, utilizes high-resolution references to enhance restoration. Work in this field, such as that by \citet{zhang2019image}, employed feature transfer via Patch Match \cite{barnes2009patchmatch} and VGG-based extractors \cite{simonyan2014very}. \citet{yang2020learning} also proposed methods that incorporate attention mechanisms for feature fusion, resulting in multi-scale feature enhancement. However, aligning reference images to target images can be error-prone, a challenge that \citet{wang2021dual} addressed with their aligned attention method, which preserved high-frequency details through spatial alignment.

The use of reference images extends beyond super-resolution to tasks such as deblurring. \citet{xiang2020deep} leveraged the sharpness from reference videos to enhance video deblurring networks, while \cite{li2022reference13, li2022deep14} refined feature matching via selection and ranking modules. \citet{zhao2022d2hnet} proposed enhancing blurry images with short-exposure guidance for blur removal, but relied on paired datasets. \citet{lai2022face} employed dual cameras to enhance blurry face images, focusing on a specific niche where the reference and source images are closely aligned through face segmentation. In contrast, our approach addresses more complex scenes with significant motion diversity and minimal overlap. \citet{liu2023reference} divided the task into single-image deblurring and reference transfer, but their method falls behind current single-image deblurring approaches. \citet{zou2023reference} introduced another solution building on previous work, yet it still struggles to effectively extract similarities between blurry and sharp domains.

Most notably, previous research has used reference images primarily to enhance supervised models, which still require paired ground truth data for training. In contrast, our work pioneers the use of reference images within a fully unsupervised framework, eliminating the need for paired datasets. By effectively leveraging reference inputs, we introduce a novel module that outperforms existing unsupervised single-image deblurring methods, enabling high-quality restoration without relying on paired data.

\myheading{Dense matching models.}
Dense flow regression methods, typically used for predicting correspondence maps in optical flow tasks, have recently been adapted for geometric matching to accommodate substantial variations in geometry and appearance. Notable among these are DGC-Net~\cite{melekhov2019dgc}, which employs a coarse-to-fine CNN framework, and GLU-Net~\cite{truong2020glu}, which overcomes the resolution constraints of previous models by using global and local correlation layers, complemented by the GOCor module \cite{truong2020gocor} for enhanced matching precision. Our method innovates within this landscape by being the first to apply Dense Matching for matching across domains based on the semantic similarity of objects, breaking new ground in the field of domain-specific dense correspondence.
\vspace{-2mm}

%% file: sec/3_method.tex
\section{BluRef -- Reference-based Deblurring}
\label{sec:method}

We approach the problem of image deblurring through an iterative enhancement process. Our goal is to estimate a sharp image from a blurry image \( I_{\text{blur}} \) of a scene by utilizing a set of \( N \) reference sharp images \( \{I_{\text{ref}}^n\}_{n=1}^N \). None of the sharp reference images needs to be the exact sharp version of the blurry input \( I_{\text{blur}} \). While the reference images should come from scenes similar to that of the blurry image, they do not need to match precisely and can be captured from different temporal and spatial perspectives. In this section, we first describe the iterative training procedure, then provide details about the self-supervised training of a dense matching model, and finally explain the pseudo-sharp image generation method. 

\vspace{-1mm}
\subsection{Iterative Target Refinement and Training }

\begin{figure}[t]
    \centering
    \includegraphics[width=0.49\textwidth]{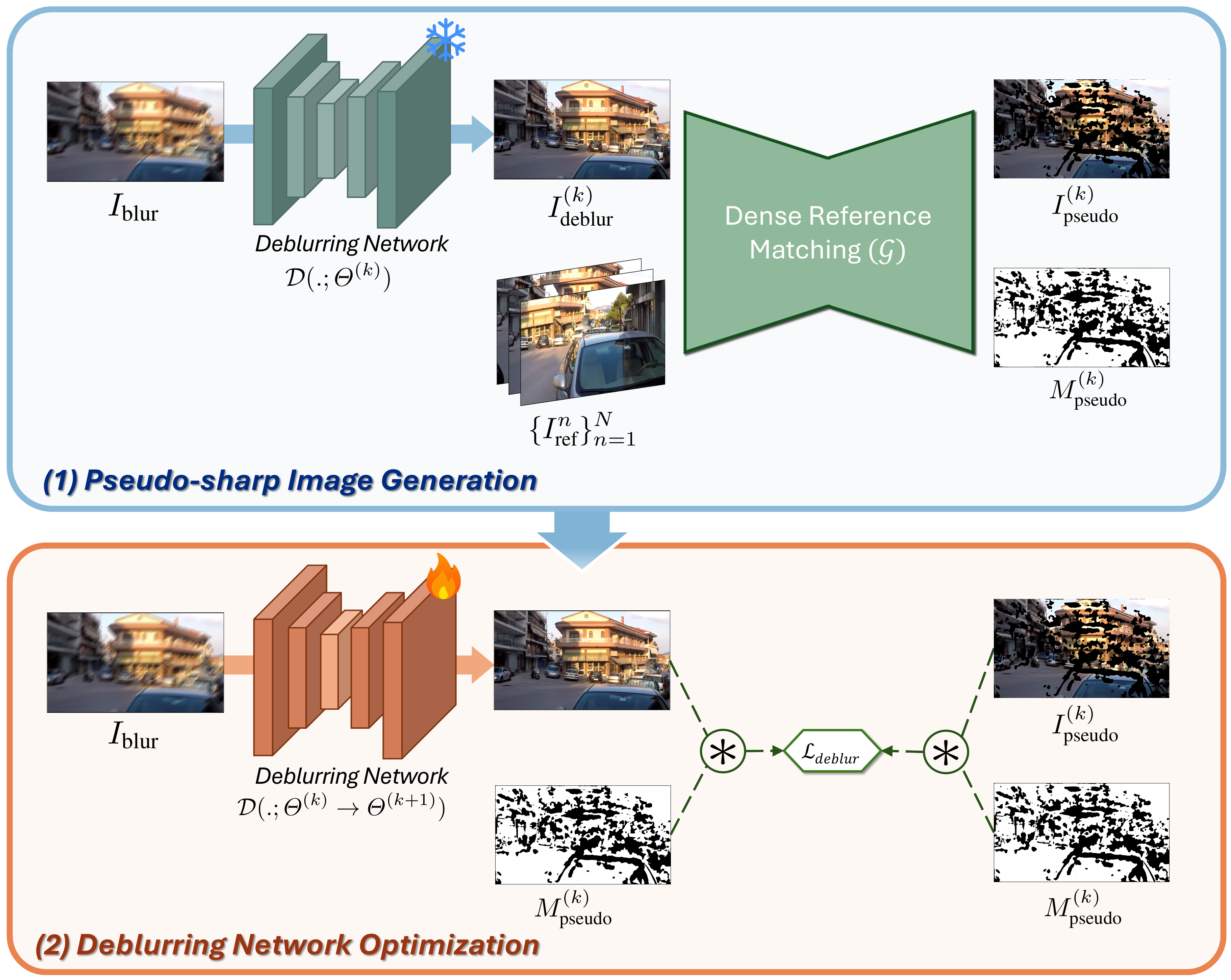}
    \caption{Iterative approach for generating a pseudo-sharp image and using it to train the deblurring model. In each epoch, the reference images \( \{I_{\text{ref}}^n\}_{n=1}^N \) are matched with the deblurred result $I_{\text{deblur}}^{(k)}$ to produce the updated pseudo-sharp image and the corresponding confidence mask, which serve as the supervision targets for training the deblurring model.}
    \label{fig:pipeline}
\end{figure}


Our iterative process comprises multiple epochs, each with two main procedures: (1) \textbf{pseudo-sharp image generation}, in which we find matching details between the blurry input and the reference sharp images to create a temporary pseudo-sharp image; and (2) \textbf{deblurring network optimization}, where we use the current pseudo-sharp image as ground truth to optimize the deblurring network backbone. This is illustrated in \Fref{fig:pipeline} and is mathematically expressed as follows:
\begin{align}
& I_{\text{deblur}}^{(k)} = \mD(I_{\text{blur}};\Theta^{(k)}), \label{eq:deblur}\\
& I_{\text{pseudo}}^{(k)}, M_{\text{pseudo}}^{(k)} = \mG(I_{\text{deblur}}^{(k)},\{I_{\text{ref}}^n\}_{n=1}^N), \label{eq:pseudo}\\
& \Theta^{(k+1)} = \myargmin{\Theta}\mathcal{L_\text{deblur}}\big(\mathcal{D}(I_{\text{blur}};\Theta); I_{\text{pseudo}}^{(k)},\, M_{\text{pseudo}}^{(k)}\big) \label{eq:lossk}
\end{align}




In the equations above, $\mD$ denotes the deblurring network parameterized by $\Theta^{(k)}$ at the $k$-th epoch, which produces a deblurred estimate $I_{\text{deblur}}^{(k)}$ from the blurry input $I_{\text{blur}}$. The function $\mG$ represents a process that takes this deblurred image and matches it with the set of unpaired sharp references $\{I_{\text{ref}}^n\}_{n=1}^N$ using a trained Dense Matching ($\mathcal{DM}$) model to generate a pseudo-sharp image $I_{\text{pseudo}}^{(k)}$ and a confidence mask $M_{\text{pseudo}}^{(k)}$ that suppresses unreliable correspondences (more details in \Sref{sec:pseudosharp}). We initialize the process with $I_{\text{deblur}}^{(0)} = I_{\text{blur}}$ to provide a stable starting point for matching, before the deblurring network learns meaningful features. The pseudo-sharp image serves as the target, and the deblurring network parameters are updated by minimizing the masked reconstruction loss $\mathcal{L}_{\text{deblur}}$, where the mask $M_{\text{pseudo}}^{(k)}$ weights the valid regions during optimization. This iterative refinement leverages the progressively enhanced deblurring model to incrementally improve the pseudo-sharp target, ensuring that, after each epoch, it more closely approximates the true sharp image.

\vspace{-1mm}

\subsection{Dense Matching -- Self-supervised Training}
\label{sec:dm}


To bridge the gap between blurred and sharp images, our strategy is to extract corresponding regions across both domains. This involves training a $\mathcal{DM}$ model to recognize patterns and similarities between blurred and sharp images, which is crucial for generating what will be referred to as a `pseudo-sharp' image. In this work, the $\mathcal{DM}$ model is defined as a function that establishes correspondences between a target image, which can be blurry or of low resolution, and a series of sharp reference images. This function is formulated as follows:
$\mathcal{DM} : (I_{\text{target}}, I_{\text{ref}}) \rightarrow (I_{\text{trans}}, M_{\text{conf}}).$
Here, \(I_{\text{target}}\) is the image we want to deblur, \(I_{\text{ref}}\) is a sharp reference, \(I_{\text{trans}}\) is the result of applying the inferred correspondences from \(I_{\text{ref}}\) to \(I_{\text{target}}\), and \(M_{\text{conf}}\) is a binary confidence mask that specifies the certainty of each transformation applied.

To train the $\mathcal{DM}$ model to effectively bridge the feature gap between blurred and sharp domains, we adopt a self-supervised scheme inspired by prior works \cite{truong2023pdc, truong2020glu, truong2020gocor, rocco2017convolutional}. This approach generates training pairs, denoted as $(I_{\text{warped}}, I_{\text{gt}})$, by applying artificial warping to an initial set of sharp images from any domain. Specifically, we create each synthetic pair by sampling a base sharp image and applying random geometric transformations, such as Homography or Thin-Plate Spline, to produce a deformed version. For producing each synthetic image pair, we first resize a given original sharp image to a larger resolution, \(H' {\times} W'\), and apply a dense flow field of the same dimensions using random geometric transformations to create a deformed image. The warped image, $I_{\text{warped}}$, is then produced by centrally cropping this deformed image to the desired resolution, \(H {\times} W\), while the ground-truth image, $I_{\text{gt}}$, is obtained by similarly cropping the original sharp image to the same \(H {\times} W\) size. To enhance the model's adaptability to match detail between sharp and blurry domains, we further augment the warped images $I_{\text{warped}}$ by blur augmentation with random blur and noise, inspired by BSRGAN~\cite{zhang2021designing}, ensuring the model trained on such data can effectively learn to match features across diverse visual conditions (more detail in Supplementary). In the context of our BluRef framework, $I_{\text{warped}}$ serves as the target image $I_{\text{target}}$, while $I_{\text{gt}}$ from training corresponds to the reference image $I_{\text{ref}}$.

\vspace{-1mm}


\subsection{Pseudo-sharp Image Generation} \label{sec:pseudosharp}

After acquiring a robust $\mathcal{DM}$ model that can identify pixel correspondences between sharp and blur images, we employ it within a pseudo-sharp generation process, denoted by $\mG$. This process produces a pseudo-sharp image \(I_{\text{pseudo}}\) for each blur image \(I_{\text{blur}}\), given the set of reference sharp images \( \{I_{\text{ref}}^n\}_{n=1}^N \). When the number of reference images $N$ is greater than one, $\mG$ aggregates multiple reference matches to form a single pseudo-sharp image. We consider three aggregation strategies: \textit{Weighted Average}, \textit{Sequential Accumulation}, and \textit{Progressive Reference Averaging}.





\myheading{Weighted Average.}
This strategy applies the $\mathcal{DM}$ model independently to each pair \( \{I_{\text{blur}}, I_{\text{ref}}^n\} \), generating a series of pseudo-sharp images \( I_{\text{pseudo}}^n \) and corresponding confidence masks \( M_{\text{pseudo}}^n \). The final \( I_{\text{pseudo}} \) and \( M_{\text{pseudo}} \) are obtained by averaging these intermediate results:
\vspace{-2mm}
\begin{align}
I_{\text{pseudo}}, M_{\text{pseudo}} = \frac{1}{N} \sum_{n=1}^{N} I_{\text{pseudo}}^n * M_{\text{pseudo}}^n, \frac{1}{N} \sum_{n=1}^{N} M_{\text{pseudo}}^n. \nonumber 
\end{align}

\myheading{Sequential Accumulation.}
This strategy iteratively refines \( I_{\text{pseudo}} \) by using the output from the previous iteration as input for the next, ensuring the continuity of sharp details:
\begin{align}
    &I_{\text{pseudo}}^n, M_{\text{pseudo}}^n =  \\ 
     &\hspace{3ex}\mathcal{DM}(I_{\text{pseudo}}^{n-1} * M_{\text{pseudo}}^{n-1} + I_{\text{blur}} * (1 - M_{\text{pseudo}}^{n-1}), I_{\text{ref}}^n), \nonumber \\
&    I_{\text{pseudo}}, M_{\text{pseudo}} = I_{\text{pseudo}}^N, M_{\text{pseudo}}^N. \nonumber 
\end{align}
By employing this method, we maintain consistency in pixel quality between iterations as \(I_{\text{pseudo}}^n\) builds upon the sharpness from \(I_{\text{pseudo}}^{n-1}\), creating a continuity of detail.

\myheading{Progressive Reference Averaging.}  
This strategy combines the advantages of previous strategies by retaining sharp details from prior iterations and selectively enhancing unmatched areas in \( I^n_{\text{pseudo}} \):
\begin{align}
    & I_{\text{pseudo}}^n, M_{\text{pseudo}}^n = \mathcal{DM}(I_{\text{blur}} * (1 - M_{\text{pseudo}}^{n-1}), I_{\text{ref}}^n), \\
    & I_{\text{pseudo}}, M_{\text{pseudo}} = \frac{1}{N} \sum_{n=1}^{N} M_{\text{pseudo}}^n * I_{\text{pseudo}}^n, \frac{1}{N} \sum_{n=1}^{N} M_{\text{pseudo}}^n. \nonumber 
\end{align}
This strategy ensures that with each iteration, newly matched details are integrated without discarding the sharpness achieved from previous matches, culminating in a high-quality pseudo-sharp image.

\subsection{Loss Function}
At each epoch, our deblurring network is trained using the common image deblurring training objectives, with the main component being a reconstruction loss. However, since not all regions in the pseudo-sharp image are predicted with high certainty, we use the confidence map \( M_{\text{pseudo}} \)  to weigh the contribution of each image region in the loss computation:
{\small
\begin{align}
\Theta^{(k + 1)} := \myargmin{\Theta} \mathcal{L}\left( \mathcal{D}(I_{\text{blur}}; \Theta) * \Bar{M}^{(k)}_{\text{pseudo}},  I^{(k)}_{\text{pseudo}} * \Bar{M}^{(k)}_{\text{pseudo}} \right). \nonumber
\end{align}
}
Here, \( \mathcal{L} \) is any suitable loss metric, including $L_1$, $L_2$, or PSNR. It can be chosen based on the specific requirements for the deblurring model. $\bar{M}^{(k)}_{\text{pseudo}}$ is the binary confidence mask, obtained by binarizing the confidence map $M^{(k)}_{\text{pseudo}}$ (using a threshold of 0.7 in our experiments).

\subsection{Inference}


At test time, BluRef operates as a direct deblurring model. Both the dense matching module and the pseudo-sharp generation process are discarded after training. Given a trained deblurring network with parameters $\Theta^\star$, the final sharp image is produced by a single forward pass:
\begin{align}
    \hat I_{\text{sharp}} = \mD(I_{\text{blur}};\Theta^\star). \nonumber
\end{align}
This means BluRef requires only a single blurry input at inference and runs with the same computational cost as a standard deblurring backbone. Despite its powerful training framework, the inference pipeline remains simple and efficient—no extra components, no multi-stage processing, just fast and direct deblurring.



%% file: sec/4_exp.tex
\section{Experiments}
\label{sec:exp}

\subsection{Experimental Setups}
\myheading{Datasets.} We use three datasets in our experiments, namely GoPro~\cite{Nah_2017_CVPR}, RB2V~\cite{pham2023hypercut}, and PhoneCraft \cite{pham2024blur2blur}. GoPro is a common image debluring dataset that consists of paired sharp and synthetic blur images. It contains 2103 training pairs across 22 scenes and 1111 test pairs from 11 scenes. RB2V, in contrast, is a collection of real blur images from varied scenes. We specifically use the RB2V\_Street subset to capture a wide range of scene types, which covers 11k training pairs and 2053 test pairs. PhoneCraft is an unpaired real-world dataset recorded with a smartphone, comprising 12 blurry and 11 sharp video clips. Each clip ranges from 30 to 40 seconds and is captured at a frame rate of 60 fps.

\begin{figure}[!htbp]
    \centering
    \includegraphics[width=0.9\linewidth]{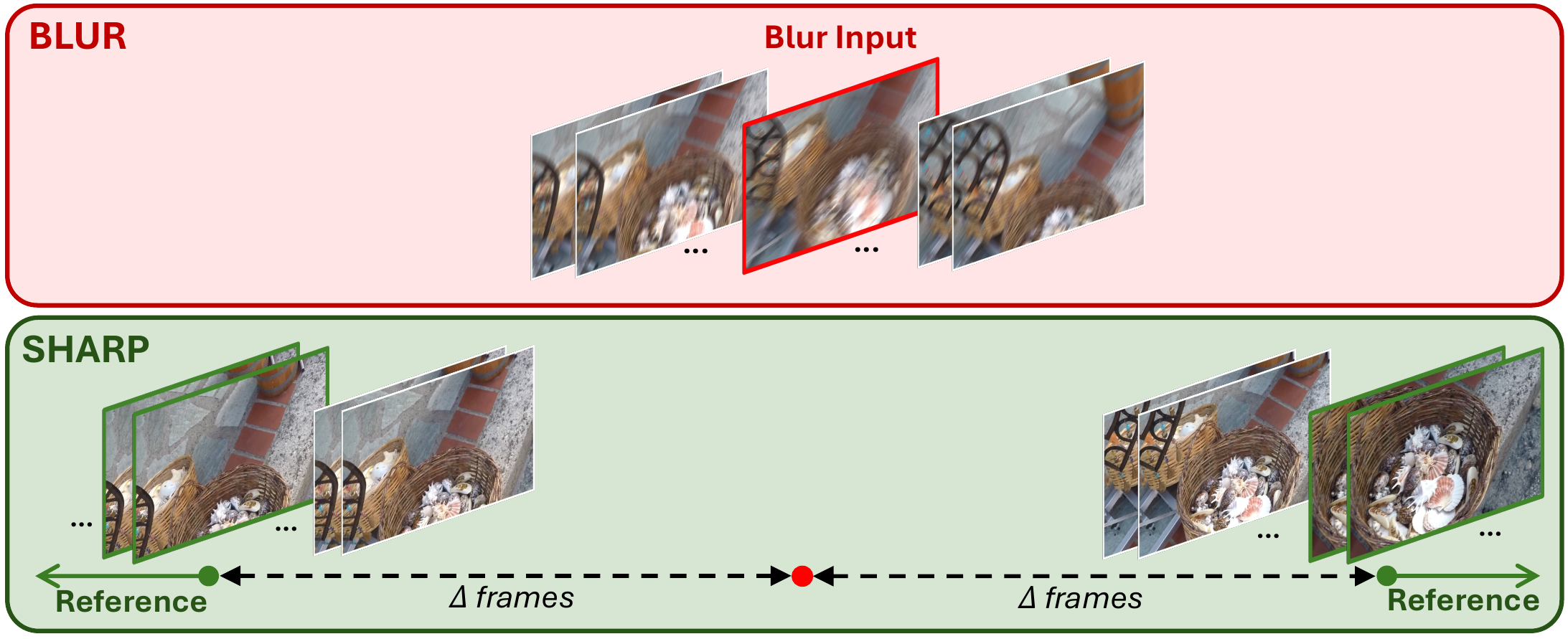}
    \caption{Protocol to collect sharp reference images in our experiments. For each blurry image, we collect \( N \) sharp reference images as the two groups of consecutive frames from both sides, each displaced by \( \Delta \) frames from the blurry image. } 
    \label{fig:dataset}
\end{figure}

To adapt GoPro and RB2V datasets to our problem, we divide each dataset into separate sets for blurry and sharp reference images, ensuring distinct indices to maintain an unsupervised learning environment. For each blurry image, we pick $N$ sharp reference images, which consist of two sets of consecutive frames shifted $\Delta$ frames from the blurry ones on both sides, as illustrated in \cref{fig:dataset}. $\Delta$ is not just a measure of temporal distance but also reflects the real-world occurrence of capturing videos with both sharp and blurry frames or, in the context of dataset preparation for deblurring, capturing a scene with cameras operating at different speeds. Then, for evaluating the deblurring performance under various temporal separations, $\Delta$ is set to values of 1, 10, and 20. Next, the number of sharp reference frames $N$ is another important hyperparameter that impacts the quality of the deblurring. We will discuss its effect in \cref{sec:ablation}. By default, we use $N {=} 6$ for all experiments in the other sections. Thus, in our experiment, the training dataset for GoPro includes 1050 blurry images alongside 1053 sharp images, whereas, for RB2V, it consists of 5400 blurry and 5600 sharp images.

Unlike GoPro and RB2V, PhoneCraft consists of separate blurry and sharp video clips recorded under different motion states. Blurry frames arise from rapid camera or object motion, whereas sharp clips are captured when the camera remains static or moves slowly in a static scene. Since no temporal alignment exists between the two, we randomly select sharp references for each blurry input to eliminate temporal correlation. The PhoneCraft dataset serves as a challenging real-world benchmark. All videos are captured by a handheld smartphone in unconstrained environments, resulting in complex blur patterns and significant temporal and geometric misalignment, with no paired sharp ground truth available. This setting closely reflects practical deployment conditions. We further analyze the performance of BluRef and existing baselines on this dataset in \cref{phonecraft}.

\myheading{Implementation Details.} All model components are trained using Adam optimizer \cite{kingma2014adam}, with $\beta_1 = 0.9$ and $\beta_2 = 0.999$.  We employ an initial learning rate of $2 {\times} 10^{-4}$ and implement a cosine annealing \cite{loshchilov2016sgdr} schedule to gradually reduce the learning rate down to $10^{-6}$ with a total of 400K iterations. During training, we randomly crop these images to obtain a square shape of $256 {\times} 256$ pixels, and we use an identical cropping box for the blurry and its reference images. For data augmentation, we apply random flips along both the horizontal and vertical axes, with a probability of 0.5. In addition, the $\mathcal{DM}$ model is built upon the PDC-Net+\cite{truong2023pdc} framework, with the integration of GLU-Net-GOCor~\cite{truong2020glu, truong2020gocor} as its base architecture. For the GOCor modules \cite{truong2020gocor}, the training encompasses three iterations of both local and global optimizations. For training the $\mathcal{DM}$ model, we use only the sharp image set from the GoPro dataset, applying random degradations inspired by BSRGAN~\cite{zhang2021designing} and adapted for blind image deblurring with motion blur augmentation to create $I_{\text{warped}}$. This prevents blur patterns from the original dataset from leaking into training, ensuring the integrity of our unsupervised framework. We use this pre-trained $\mathcal{DM}$ model for all BluRef experiments in this paper. Lastly, we use the network architectures of NAFNet~\cite{nafnet} and Restormer~\cite{zamir2022restormer} as the backbone of our deblurring model $\mD$, which is trained from scratch using BluRef.

%

\myheading{Baselines.} We evaluate BluRef's effectiveness with a comprehensive list of baseline methods, encompassing both unsupervised (DualGAN\cite{yi2017dualgan}, UID-GAN~\cite{lu2019uid}, UAUD~\cite{tang2023uncertainty}) and supervised approaches (NAFNet \cite{nafnet}, Restormer \cite{zamir2022restormer}) on RB2V and GoPro dataset. Specifically, for the BluRef framework, we assessed the performance across its three distinct strategies: Weighted Average ({\bf Avg.}), Sequential Accumulation ({\bf Seq.}), and Progressive Reference Averaging ({\bf Prog.}), each integrated with the aforementioned supervised techniques. To assess performance in fully unpaired real-world scenarios, we evaluate reblurring-based adaptation methods (Blur2Blur~\cite{pham2024blur2blur}, BSRGAN~\cite{zhang2021designing}) on the PhoneCraft dataset.

To ensure fair comparisons, all models were trained under unpaired data configurations segmented according to the $\Delta$ value. Notably, the results from the supervised models serve as upper bound benchmark, as they are trained on paired blur-sharp image pairs from each subset.

\subsection{Experimental Results}
\myheading{Quantitative Comparison.} As shown in \cref{tab:quan}, existing unsupervised methods perform significantly worse than BluRef on both synthetic and real-world datasets. In contrast, BluRef—when paired with strong backbones such as NAFNet or Restormer—achieves performance close to, and in some settings even beyond, supervised models. Notably, our proposed method with the NAFNet backbone and the Progressive Averaging strategy achieves a PSNR of 31.94 dB on GoPro for $\Delta = 1$ frame, which is close to the supervised NAFNet's score of 33.32 dB. 

On the RB2V dataset, our method with Restormer and the Progressive Averaging strategy outperforms the supervised Restormer's upper bound of 27.43 dB with a PSNR of 27.87 dB and 27.72 dB for $\Delta = $ 1 frame and 10 frames, respectively, showcasing the potential of our approach to exceed supervised methods in conditions where we can leverage information from surrounding frames. Results for the Weighted Average and Sequential Accumulation strategies follow similar trends with NAFNet and are included in the supplementary material.

When increasing the temporal distance \( \Delta \) from 1 to 20 frames, BlurRef only shows a minor decrease in  PSNR and SSIM scores on both datasets and with both deblurring backbones. This confirms BluRef’s robustness when reference frames differ significantly from the blurry input due to motion or content changes. The ability to remain stable under large $\Delta$ values highlights BluRef’s practicality for constructing pseudo pairs from real-world, unaligned video data.

\begin{table*}[t]
\centering
\aboverulesep=0ex 
\belowrulesep=0ex 

\resizebox{0.83\linewidth}{!}{%
\setlength{\tabcolsep}{5pt}
\begin{tabular}{l|ccc|ccc}
\toprule\rule{0pt}{1.1EM}
             & \multicolumn{3}{c|}{\textbf{GoPro}}    & \multicolumn{3}{c}{\textbf{RB2V}} \\
\midrule\rule{0pt}{1.1EM}
\textbf{Delta ($\Delta$)}         & \textit{1 frame}       & \textit{10 frames}      & \textit{20 frames}       & \textit{1 frame}       & \textit{10 frames}       & \textit{20 frames}     \\ 
\midrule\rule{0pt}{1.1EM}
\textbf{Unsupervised Deblurring} & & & & & & \\
\quad DualGAN & \metric{22.23/0.721} & \metric{22.10/0.719} & \metric{21.24/0.702} & \metric{21.01/0.512} & \metric{20.87/0.500} & \metric{20.92/0.505}  \\
\quad UID-GAN & \metric{23.42/0.732} & \metric{23.18/0.724} & \metric{22.38/0.724} & \metric{22.22/0.578} & \metric{22.01/0.551} & \metric{22.13/0.569}  \\
\quad UAUD & \metric{24.25/0.792} & \metric{24.02/0.750} & \metric{23.77/0.745} & \metric{22.87/0.590} & \metric{22.29/0.581} & \metric{22.28/0.581}  \\
\quad \textbf{BluRef (Ours)} & & & & & & \\

\quad \quad NAFNet - BluRef (Avg.) & \metric{29.32/0.933} & \metric{29.21/0.915} & \metric{29.15/0.911} & \metric{25.97/0.783}& \metric{25.96/0.783} & \metric{25.65/0.775}\\
\quad \quad NAFNet - BluRef (Seq.) & \metric{29.82/0.947} & \metric{29.68/0.940} & \metric{29.60/0.940} & \metric{26.14/0.790} & \metric{26.02/0.787} & \metric{25.93/0.780}  \\
\quad \quad NAFNet - BluRef (Prog.) & \textbf{\metric{31.94/0.960}} & \textbf{\metric{31.87/0.955}} & \textbf{\metric{31.52/0.947}} & \textbf{\metric{27.87/0.821}} & \textbf{\metric{27.72/0.820}} & \textbf{\metric{27.24/0.812}}  \\
\quad \quad Restormer - BluRef (Prog.) & \metricu{31.02/0.950} & \metricu{30.97/0.949} & \metricu{30.95/0.938} & \metricu{26.82/0.839} & \metricu{26.76/0.832} & \metricu{26.13/0.829} \\
\midrule
\textit{\textbf{Supervised - Upperbound}} & & & & & & \\
\quad NAFNet     & \multicolumn{3}{c|}{\metric{33.32/0.962}} & \multicolumn{3}{c}{\metric{28.54/0.824}}  \\
\quad Restormer  & \multicolumn{3}{c|}{\metric{32.92/0.961}} & \multicolumn{3}{c}{\metric{27.43/0.849}}  \\
\bottomrule
\end{tabular}
}
\caption{Comparison of different deblurring methods on GoPro and RB2V datasets. For each test, we report \psnr{PSNR$\uparrow$}/\ssim{SSIM$\uparrow$} scores as evaluation metrics. The best scores are in \textbf{bold} and the second best score are in \underline{underline}. For a supervised method, NAFNet or Restormer, we assess its upperbound of deblurring performance by training it on the paired dataset.  \label{tab:quan}
}
\vspace{-4mm}
\end{table*}

\begin{figure*}[!htb]
    \centering
    \includegraphics[width=0.92\textwidth]{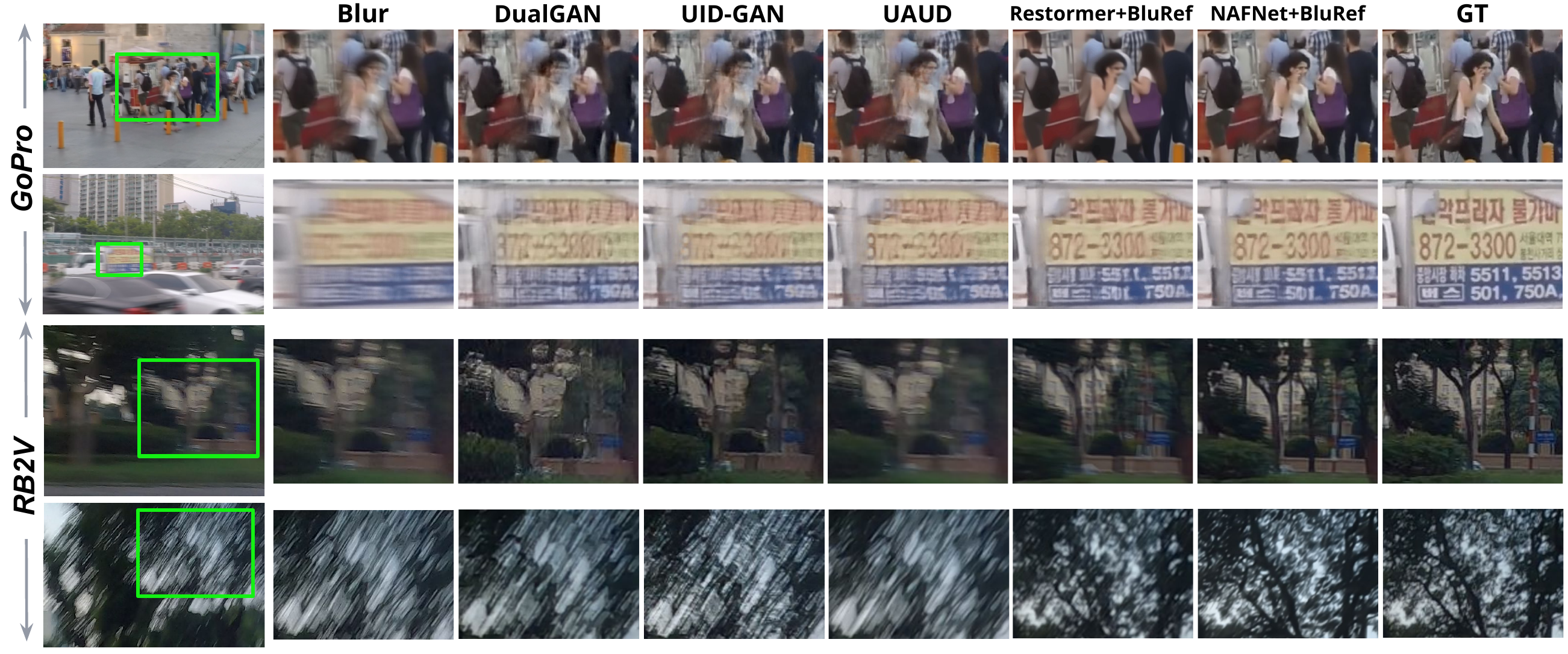}
    \caption{Qualitative results of several methods on GoPro and RB2V data. This figure presents four examples, two from each dataset. For each example, a zoomed-in portion of the input image is displayed, accompanied by its ground truth counterpart and the results from BluRef and three other baselines. Additional results are included in the supplementary material.}
    \label{fig:qual}
    \vspace{-4mm}
\end{figure*}

\myheading{Qualitative results.} We provide a qualitative comparison between our method and baseline methods in \cref{fig:qual}. As can be seen, the unsupervised approaches, including DualGAN, UID-GAN, and UAUD, produce poor results on both datasets. Particularly, they fail to recover the lady's body in the first example and retain strong motion blur artifacts in the second one. In contrast, BluRef, when combined with either NAFNet or Restormer backbone, can produce deblurred images with much clearer details, close to upperbound result from NAFNet and approximating the sharp ground truth. It recovers correctly the body shape, face, and hand position of the lady in the first example and the tree branches in the second example. This result is satisfactory, given the challenging inputs with extreme motion blurs.

\subsection{Ablation Studies}
\label{sec:ablation}

\myheading{Number of referenced images.}
We investigate how the number of reference images influences deblurring performance by evaluating BluRef with 4, 6, 8, and 10 reference frames. For this study, we use a NAFNet backbone, $\Delta= 1$ and the Progressive Averaging strategy based on the superior results presented in \cref{tab:quan}. As shown in \cref{tab:numofref}, BluRef performs robustly across a wide range of reference counts, with 6 and 8 frames yielding similarly strong results - 6 achieving slightly higher PSNR, and 8 providing marginally better SSIM. Using only 4 references provides limited scene coverage and results in a mild drop, whereas using 10 references may introduce redundant or strongly misaligned content that degrades the aggregated pseudo-ground truth. Importantly, BluRef is not sensitive to the exact choice of $N$: all settings from 4 to 10 frames achieve above 31 dB PSNR, indicating that a moderate number of references is sufficient and that the method remains stable even when the reference set is not perfectly aligned. In practice, videos naturally provide many usable frames, and BluRef does not require users to manually select “good’’ references—its confidence masks automatically suppress unreliable matches.


\begin{table}[!htbp]
    \centering
\aboverulesep=0ex 
\belowrulesep=0ex 
    \resizebox{0.93\columnwidth}{!}{%
    \setlength{\tabcolsep}{3pt}
    \begin{tabular}{l|cccc}
        \toprule\rule{0pt}{1.1EM}
         Number of Refs.            & \textit{4 frames} & \textit{6 frames} & \textit{8 frames} & \textit{10 frames} \\
        \midrule\rule{0pt}{1.1EM}
        \metric{PSNR$\uparrow$/SSIM$\uparrow$}  & \metric{31.42/0.942} & \psnr{\textbf{31.94}}/\ssim{0.960} & \psnr{31.93}/\ssim{\textbf{0.961}} & \metric{31.05/0.924} \\
        \bottomrule
    \end{tabular}
    }
    \caption{Effect of the number of reference images on BlurRef performance on the GoPro dataset.     \label{tab:numofref}}
\end{table}

\begin{figure}[!htbp]
\centering
\includegraphics[width=0.75\linewidth]{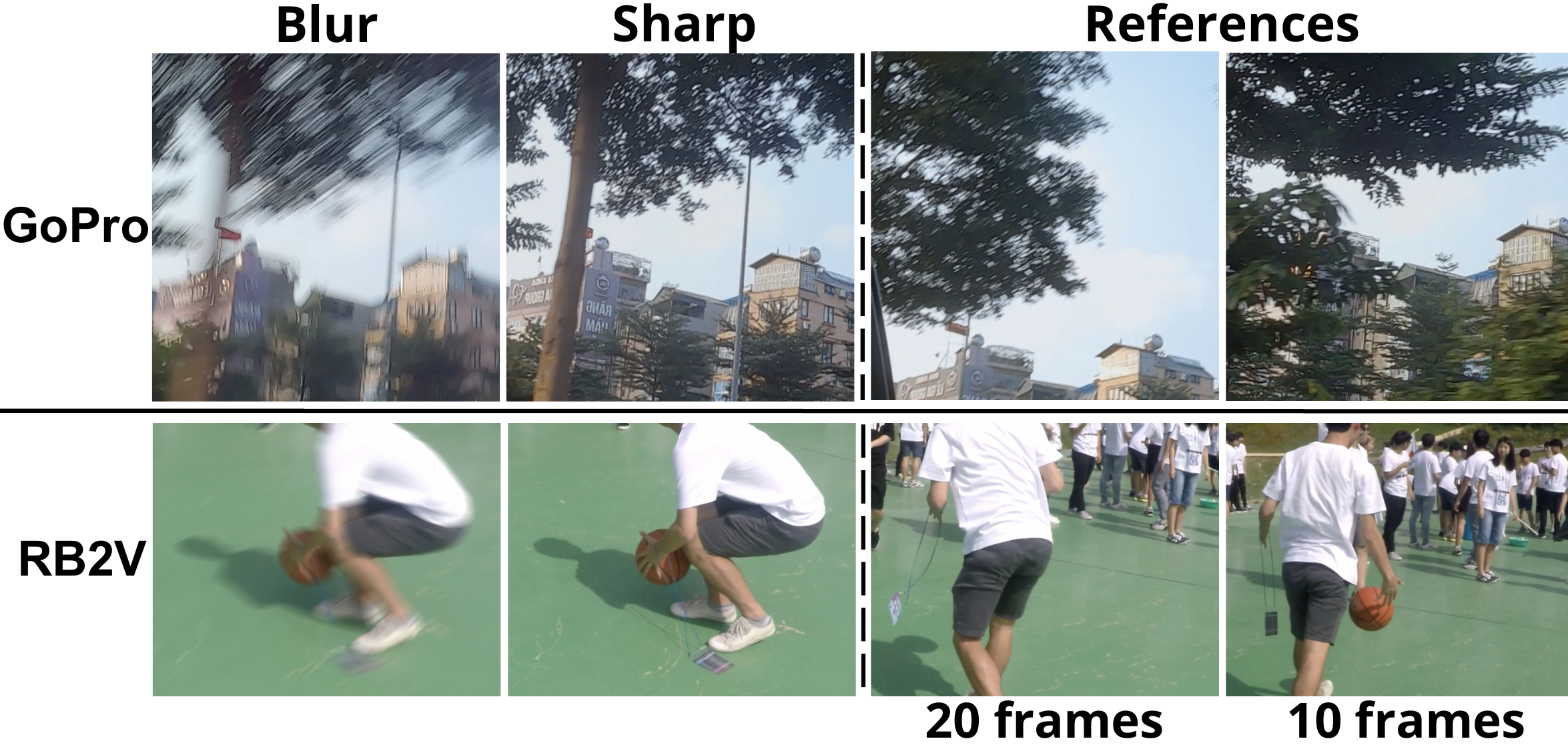}
\caption{Visualization of blur and sharp images along with corresponding reference frames used for training BluRef.}
\label{fig:delta}
\end{figure}%

\myheading{Matching content between blur and sharp.}
The datasets used in our experiments feature continuous camera and object movements; hence, frames with a small index discrepancy \(\Delta\) can exhibit significant content differences. As shown in \Fref{fig:delta}, we present examples of the substantial content gap between the training blurry image and reference images when \(\Delta = 10\) and \(\Delta = 20\). To demonstrate the challenge of our setting, we conducted experiments using PDC-Net+ to extract correlations with a threshold of 0.7. We calculated the percentage of matched regions between the ground truth and the reference images separated by \(\Delta\) frames. As reported in \cref{tab:matching}, all correlation percentages are below 40\%, indicating significant content variation. Despite this, our method performs effectively, showing only a slight difference in deblurring performance compared to when \(\Delta = 1\). 

\begin{table}
\centering
\small
\aboverulesep=0ex 
\belowrulesep=0ex 
\resizebox{0.4\columnwidth}{!}{%
\setlength{\tabcolsep}{5pt}
\begin{tabular}{l|cc}
    \toprule
    Dataset & GoPro & RB2V \\
    \midrule
    ${\Delta} = 10$  & 36.1\% & 33.7\% \\
    ${\Delta} = 20$  & 28.4\% & 25.2\% \\
    \bottomrule
\end{tabular}}
\caption{Percentage of matching content between blur and sharp images across the dataset. }
\label{tab:matching}
\end{table}

\vspace{-1mm}
\subsection{Evaluating the Real-World Scenarios}
\label{phonecraft}

\myheading{Unpaired real-blur dataset -- PhoneCraft.} 
We further examine BluRef's practical impact using the PhoneCraft dataset, which comprises videos recorded by a single moving camera across everyday scenarios. This setup lacks paired ground truth, posing a challenge for supervised deblurring while providing a rigorous test environment for unsupervised approaches. 
For BluRef training, we utilize blurred and sharp videos from the PhoneCraft dataset that capture the same scene but are temporally and geometrically misaligned. For baselines, we select BSRGAN as generalized image deblurring model and the state-of-the-art unsupervised deblurring method Blur2Blur~\cite{pham2024blur2blur}, both combined with a pretrained NAFNet backbone. Blur2Blur is an unsupervised method that transforms a blurry image into an intermediate blur domain where a pretrained deblurring network, trained on paired data in that domain, can be applied. In this context, when deblurring an image from PhoneCraft using Blur2Blur, we can choose either RSBlur~\cite{rsblur} or GoPro as the intermediate blur domain.
\begin{table}[!htbp]
    \centering 
    
    \setlength{\tabcolsep}{5pt}
    \resizebox{0.93\columnwidth}{!}{%
    \begin{tabular}{lc}
        \toprule
      Models   & NIQE$\downarrow$/FID$\downarrow$\\
      \midrule
      BSRGAN   &  13.34/10.25 \\
      Blur2Blur (GoPro as intermediate blur)   & 12.01/8.93\\
      Blur2Blur (RSBlur as intermediate blur)  & \underline{10.07}/\underline{6.28}\\
      \midrule
      BluRef & 10.43/6.45 \\
      BluRef + Blur2Blur (RSBlur as intermediate blur)   & \textbf{8.47/5.62} \\
      \bottomrule 
    \end{tabular}
    }
    \caption{Performance comparison of unsupervised models on the PhoneCraft dataset. The best scores are in \textbf{bold} and the second best scores are \underline{underlined}. \label{tab:real1}}
\end{table}



\cref{tab:real1} reports NIQE and FID scores. Although Blur2Blur achieves strong results, it requires closely matched source and intermediate blur kernel distributions. 
This dependence becomes evident when using GoPro as the intermediate blur: its synthetic blur characteristics fail to match the real-world blur in PhoneCraft, causing a performance drop compared to the better-aligned RSBlur.
In contrast, BluRef circumvents precise domain alignment by utilizing easily obtainable reference frames, robust to significant temporal and geometrical misalignments. This allows BluRef to maintain consistently high-quality results in real-world conditions, without relying on specific blur characteristics or domain-specific adaptations.

\begin{figure*}[!htbp]
    \centering
    \includegraphics[width=0.95\linewidth]{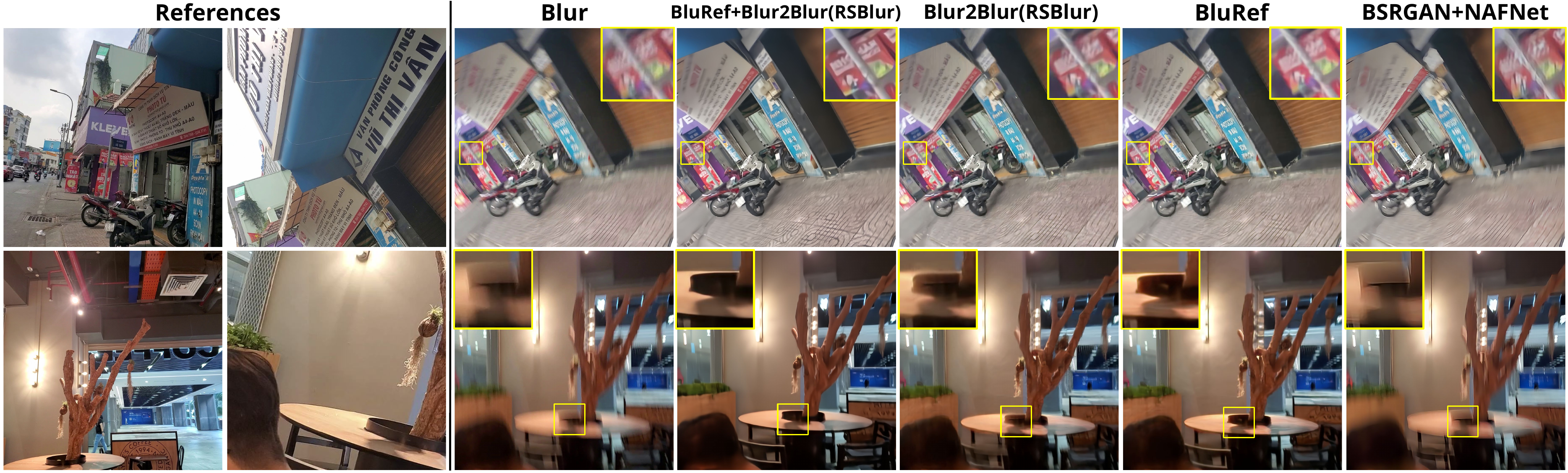}
    \caption{Qualitative comparison of BluRef, Blur2Blur, and BSRGAN methods for real-world scenario with PhoneCraft dataset.}
    \label{fig:phonecraft}
\end{figure*}
\begin{figure}[!htbp]
    \centering
    \includegraphics[width=0.8\linewidth]{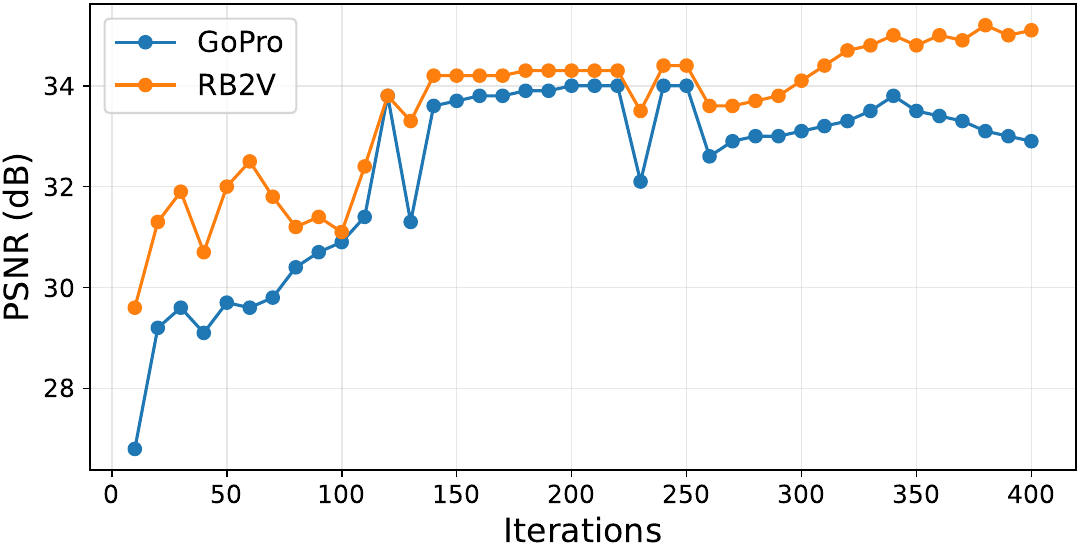}
    \caption{PSNR scores (by training iteration) of pseudo-sharp images generated by BluRef when comparing to the ground-truths.}
    \label{fig:psnr}
\end{figure}
\myheading{Challenging the supervised paradigm.}
Although Blur2Blur faces limitations when the blur kernels of the source and target datasets differ significantly, it provides a key advantage in transferring blur kernel characteristics between domains. We can leverage this advatange by combining BluRef with Blur2Blur to enhance the performance of both. Specifically, Blur2Blur first adapts the intermediate blur dataset, RSBlur, aligning its blur characteristics with those of PhoneCraft. Instead of training BluRef's deblurring model from scratch, we initialize it with a pretrained NAFNet on RSBlur.

This approach allows Blur2Blur to reduce the domain gap initially, while BluRef's iterative refinement further enhances the model's robustness, enabling it to handle a broader range of blurs effectively. As can be seen in \cref{tab:real1}, this combined approach significantly improves performance when deblurring previously unseen blurs, such as those in PhoneCraft. \Fref{fig:phonecraft} presents the qualitative results of the combined approach, along with the individual contributions of each model.

On the GoPro and RB2V datasets, where ground truth references are available, we use PSNR and SSIM metrics to compare the combined BluRef+Blur2Blur method against the supervisedly trained NAFNet model. In this experiment, for Blur2Blur, we use GoPro and RB2V as the source domains and RSBlur as the intermediate blur domain. For BluRef training, we set the temporal offset to $\Delta = 10$ to capture challenging reference frames that reflect realistic conditions. The results in \cref{tab:real2} show that Blur2Blur-BluRef improves robustness and outperforms the supervised method on RB2V, a real-world blur dataset. 
\begin{table}[!htb]
    \centering
    \setlength{\tabcolsep}{5pt}
    \resizebox{0.9\columnwidth}{!}{
    \begin{tabular}{lcc}
    \toprule 
         & GoPro & RB2V\\
         \midrule 
        BluRef & \metric{31.87/0.955} & \metric{27.72/0.808} \\
        Blur2Blur (RSBlur) &  \metric{31.20/0.932} & \metric{27.97/0.812} \\
        BluRef + Blur2Blur (RSBlur) & \psnr{33.30} /  \ssim{\textbf{0.963}}  & \psnr{\textbf{29.62}} / \ssim{\textbf{0.872}} \\
        \midrule
        Supervised - \textit{uppperbound} & \psnr{\textbf{33.32}} /  \ssim{0.962} & \psnr{28.54} / \ssim{0.824} \\
        \bottomrule
    \end{tabular}
    }
    \caption{Comparison performance between combining BluRef and Blur2Blur with supervised deblurring network. For each test, we report \psnr{PSNR$\uparrow$}/\ssim{SSIM$\uparrow$} scores as evaluation metrics.  }
     \label{tab:real2}

\end{table}
This demonstrates that unsupervised models can perform effectively in real-world conditions, even surpassing a supervised method that requires difficult-to-collect paired training data.

\subsection{Paired Dataset Building -- Potential Benefits} 
\label{sec:benefits}
A useful byproduct of BluRef is the set of pseudo–ground-truth images generated during iterative refinement. Paired with their corresponding blurry inputs, these pseudo pairs can serve as training data for deblurring networks of various capacities, including lightweight models. Below, we evaluate the quality and utility of these pseudo pairs.


\myheading{PSNR of psuedo-sharp images.}
We compute the PSNR between pseudo-sharp images and the ground truth over masked high-confidence regions on RB2V and GoPro (\cref{fig:psnr}). Initially, the PSNR is low as the deblurring model is still learning the blur kernel, causing dense matching to struggle. However, after ~100K iterations, the pseudo-sharp images clearly converge towards the ground truth, demonstrating progressive refinement. Visualizations are in the supplementary material

\begin{table}[t]
    \centering
\aboverulesep=0ex 
\belowrulesep=0ex 
    
    \resizebox{0.9\columnwidth}{!}{%
    \setlength{\tabcolsep}{5pt}
    \begin{tabular}{l|cc}
        \toprule
         Annotation for training            & GoPro & RB2V \\
        \midrule
        Real ground truth & \metric{33.32/0.962} & \metric{28.54/0.824}\\
        BluRef pseudo-ground truth & \metric{31.97/0.949} & \metric{27.73/0.817}  \\
        \bottomrule
    \end{tabular}
    }
    \caption{Comparison between the models trained on original ground-truth data (upperbound) and the models trained on paired datasets generated by BluRef. For each test, we report \psnr{PSNR$\uparrow$}/\ssim{SSIM$\uparrow$} scores as evaluation metrics.     \label{tab:app1}}
\end{table}
\myheading{Training with generated paired data.} To assess the efficacy of our pseudo-sharp images in constructing a paired dataset for image deblurring training, we trained a NAFNet~\cite{nafnet} model on this new dataset and compared it with the model trained on the original ground truth. We set $\Delta =10$ and use 6 reference images. The results, as shown in \cref{tab:app1}, indicate that the deblurring effectiveness achieved using our pseudo-sharp images closely matches that of the original dataset, with a PSNR discrepancy $< 1$~dB in RB2V.


\begin{table}[t]
\centering
\aboverulesep=0ex 
\belowrulesep=0ex 

\resizebox{0.9\columnwidth}{!}{%
\setlength{\tabcolsep}{3pt}
\begin{tabular}{ccc}
\toprule\rule{0pt}{1.1EM}
 NAFNet (light) & BSRGAN + NAFNet & RSBlur + NAFNet \\
\midrule\rule{0pt}{1.1EM}
\textbf{10.25/6.85} & 13.34/10.25 & 11.4/7.87 \\
\bottomrule
\end{tabular}
}
\caption{Comparison between the model trained on the paired dataset generated by BluRef and two generalized image deblurring baselines on the PhoneCraft dataset, using NIQE$\downarrow$/FID$\downarrow$ metrics. 
\label{tab:app2}}
\end{table}

\myheading{Evaluating on `BluRef dataset'.}
We further investigate the impact of the `BluRef dataset' in realistic applications using the unpaired real-world dataset PhoneCraft \cite{pham2024blur2blur}. Since the paired ground truth is unknown, we pick two generalized image deblurring models as the baselines for comparison, including BSRGAN~\cite{zhang2021designing} and RSBlur~\cite{rsblur}. These models are based on the NAFNet64 backbone, the default configuration for the released NAFNet. For BluRef, we opt for a more compact NAFNet version with a width of 32 and reduced to 18 blocks. The model outputs are evaluated using no-reference quality metrics, NIQE and FID, which measure the perceptual quality and fidelity of deblurred outputs without requiring paired ground truth. As shown in \cref{tab:app2}, the results reveal that in unsupervised settings with real-world blurry datasets like PhoneCraft, the BluRef pipeline exhibits significant performance improvements over other data synthesis approaches. This underscores the efficacy of BluRef in generating data that can help enhance the quality of deblurred image.

%% file: sec/5_conclusion.tex
\vspace{-2mm}
\section{Conclusions}
\label{sec:conclusion}
\vspace{-1mm}
We have introduced \textbf{BluRef}, a novel unsupervised image deblurring approach that eliminates the need for paired training data by generating high-quality pseudo-ground truths through dense matching with reference images. BluRef achieves state-of-the-art performance on the GoPro and RB2V datasets, matching the quality of supervised models, and demonstrates adaptability in real-world scenarios such as the unpaired PhoneCraft dataset. Furthermore, combining BluRef with Blur2Blur enhances robustness and bridges domain gaps, achieving superior results on diverse real-world blur conditions. These findings establish BluRef as a powerful and practical solution for unsupervised image deblurring in both controlled and real-world settings.
